\documentclass[3p]{elsarticle}
\usepackage[utf8]{inputenc}
\usepackage{adjustbox}
\usepackage{amsmath}
\usepackage{dcolumn}
\usepackage{graphics}
\RequirePackage{color}
\RequirePackage{amssymb}
\RequirePackage{graphicx}
\usepackage{multirow}	
\usepackage{url}
\usepackage{booktabs}
\usepackage{enumerate}
\usepackage[caption=false,font=footnotesize]{subfig}
\usepackage{esvect}
\usepackage{pgfplots}
\usepackage{tikz}
\usepackage{subfig}
\usetikzlibrary{calc,arrows.meta,angles,bending,quotes,shapes.geometric}
\usepackage[strings]{underscore}
\usepackage{microtype}

\usepackage{algorithm}
\usepackage{algorithmic}

\newcounter{defnCounter}
\newtheorem{defn}[defnCounter]{Definition}

\usepackage{enumitem}
\usepackage{verbatim}

\newtheorem{theorem}{Theorem}

\newlength{\fourImgSCWidth}
\newcommand{\etal}[0]{\emph{et al.}}
\newcommand{\ie}{\textit{i}.\textit{e}.}
\newcommand{\eg}{\textit{e}.\textit{g}.}

\definecolor{myRed}{rgb}{.5,.05,.05}
\definecolor{myGreen}{rgb}{.05,.5,.05}
\definecolor{myBlue}{rgb}{.05,.05,.5}
\definecolor{myYellow}{rgb}{.9,.9,.2}

\newcommand\MyBox[1]{
	\fbox{\lower0.75cm
		\vbox to 1.7cm{\vfil
			\hbox to 1.7cm{\hfil\parbox{1.4cm}{\centering#1}\hfil}
			\vfil}}}

\usepackage{pifont}
\usepackage{booktabs}

\usepackage{arydshln}

\usepackage{url}

\begin{document}

\begin{frontmatter}

\title{Neuro-inspired edge feature fusion using Choquet integrals}

\author[dptoEIM,smartcities,biomed]{Cedric Marco-Detchart\corref{mycorrespondingauthor}}
\cortext[mycorrespondingauthor]{Corresponding author}
\ead{cedric.marco@unavarra.es}
\author[FURG2]{Giancarlo Lucca}
\ead{giancarlo.lucca@furg.br}
\author[dptoEIM,biomed]{Carlos Lopez-Molina}
\ead{carlos.lopez@unavarra.es}
\author[dptoEIM,smartcities,biomed]{Laura De Miguel}
\ead{laura.demiguel@unavarra.es}
\author[dptoEIM,FURG]{Gra\c{c}aliz Pereira Dimuro}
\ead{gracalizdimuro@furg.br}
\author[dptoEIM,smartcities,biomed]{Humberto Bustince}
\ead{bustince@unavarra.es}

\address[dptoEIM]{Dpto. Estadistica, Informatica y Matematicas, Universidad Publica de Navarra, Campus Arrosadia, 31006 Pamplona, Spain}
\address[smartcities]{Institute of Smart Cities, Universidad Publica de Navarra, Campus Arrosadia, 31006 Pamplona, Spain}
\address[biomed]{NavarraBiomed, Complejo Hospitalario de Navarra, 31008 Pamplona, Spain}
\address[FURG]{Centro de Ci\^{e}ncias Computacionais, Universidade Federal do Rio Grande, Av. It\'alia km 08, Campus Carreiros, Rio Grande, 96201-900, Brazil}
\address[FURG2]{Programa de Pós-Graduação em Modelagem Computacional, Universidade Federal do Rio Grande, Av. It\'alia km 08, Campus Carreiros, Rio Grande, 96201-900, Brazil}

\begin{abstract}
    It is known that the human visual system performs a
	hierarchical information process in which early vision cues (or primitives)
	are fused in the visual cortex to compose complex shapes and descriptors.
While different aspects of the process have been extensively studied,
	as the lens adaptation or the feature detection,
	some other,
	as the feature fusion,
	have been mostly left aside.
In this work we elaborate on the fusion of early vision primitives
	using generalizations of the Choquet integral, and novel aggregation operators that have
	been extensively studied in recent years.
We propose to use generalizations of the Choquet integral to sensibly fuse elementary 
	edge cues, in an attempt to model the behaviour of neurons in 
	the early visual cortex.
Our proposal leads to a full-framed edge detection algorithm,
	whose performance is put to the test in state-of-the-art boundary detection datasets. \end{abstract}


\end{frontmatter}

\section{Introduction}

The Human Visual System (HVS) has been, historically, a source of inspiration for 
	researchers in computer vision.
The advent of Convolutional Neural	Networks~\cite{Krizhevsky12}, 
	featuring far more complex and powerful schemas than initial attempts for image processing
	using neural networks~\cite{Chellappa98},
	might seem to indicate that the ultimate simulation step has been reached:
	the (yet modest) simulation of neural tissue.
However, it is well known that the human visual system does not consist of 	
	randomly connected layers of neural networks coping with the information gathered by 
	cones and rods.
Instead, it features a more evolved system in which information 
	is subsequently analysed to detect basic shapes, which are further
	combined to detect more complex structures.
Ever since the experiments by Barlow~\cite{Barlow53,Barlow61} and 	
	Hubel and Wiesel~\cite{Hubel61,Hubel62},
	it is known that the ganglion cells in the very retina use organised
	cell structures to seek for pictorial primitives.
This study has been repeated in different animals,
	with similar conclusions~\cite{Sceniak99,Dacey00}.
Specifically, humans are known to have center-surround receptive fields at the retina,	
	which are further combined at the Early Visual Cortex (V1) to compose lines, boundaries, etc.
This discovery had a massive impact in computer vision in the early 80's, and
	in fact led to the introduction of Marr-Hildreth's Laplacian-based edge detection method~\cite{Marr80,Smith88}.
This edge detection method aimed at faithfully simulating the role of retinal ganglions using
	Laplacian (or, equivalently, Difference of Gaussian) kernels~\cite{Sotak89,Forshaw88}, in an attempt to 
	simulate the early stages of recognition in the HVS.
Different theories have been developed in the context of image processing,
    trying to mimic the features of the HVS.
A relevant case is the use of techniques from Fuzzy Set Theory
    to cope with ambiguity, noise and blurring~\cite{melin2014edge,martinez2019general}.
These ideas are often combined with bio-inspired optimization techniques~\cite{gonzalez2016optimization}.
In this way the inherent uncertainty of the edge detection process is channelled, 
	bringing better performance to edge detection.
Other approaches in literature avoid the (human-directed) modelling of the 
    happenings in the HVS,
    relying instead on machine learning techniques (such as random decision forests~\cite{Dollar2013})
    to somehow simulate the neural interaction in brain tissue.
While these detectors often give very good performance and real-time detection, they do not attempt to mimic the HVS and require a training phase for model learning.

In general, (convolutional) neural networks deeply rely on the optimization of their weights,
	as well as on the presence of multiple maxima in their parameter space,
	to mimic neural plasticity and reach good computational solutions.
Some of these networks have shown excellent performance, 
    being close or even surpassing the human performance in statistical terms~\cite{xie2015holistically,poma2020dense}.
In addition, some have extreme difficulties in performing when input
    data is significantly different from training data~\cite{Nguyen14}.
However, the artificial neural network might not be neither the 
	only nor the best way to simulate the HVS,
	despite the fact that most of this system is actually supported by neural tissue.
Specifically, we 
	believe that a system based on the extraction of primitive features,
	followed by some sensible fusion process,
	might have a better rapport with how the HVS actually works.
	
\begin{figure*}
	\centering
	\includegraphics[width=0.95\textwidth]{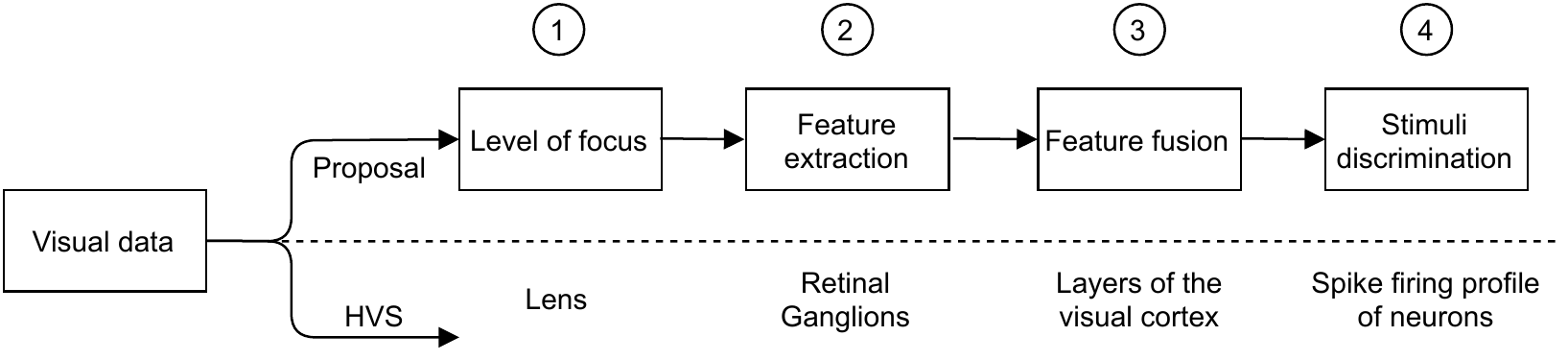}
	\caption{Schema of the multi-step algorithm proposed along with the different equivalent HVS phases.}
	\label{HVS}
\end{figure*}
	
We believe that a sensitive structure for object detection (in this case, for edge detection) 
	is a multi-step algorithm (Fig.~\ref{HVS}),
	featuring the phases visual information undergoes in the HVS.
First, visual data is to be set to the adequate level of focus,
	a task which is carried out by the lens in the HVS.
Then, some basic features are extracted from the resulting image,
	representing the role of retinal ganglions.
The third step consists of fusing the feature information to produce goal-oriented representations,
	as it happens in successive layers of the visual cortex.
Finally, the different stimuli need to be discriminated among those actually
	corresponding to the presence of an object (edge), and those to be discarded.
This thresholding process embodies the basic spike firing profile of neurons
	in human neural tissue.
The fact that this 4-phase schema becomes both computationally handy and 
	familiar~\cite{Bezdek98} is not a random coincidence.
In the early steps of computer vision, 
	computational solutions were highly influenced by advances in neuroscience and psychology.
These solutions were further adopted by researchers in upcoming decades,
	e.g. the edge detection frameworks by Bezdek et al.~\cite{Bezdek98} and Law et al.~\cite{Law96}.
	
Analysing the previous schema,
	it comes clear than some phases have been significantly more studied than others.
While image regularization (either adaptive or not) and feature extraction are well-covered in literature,
	feature fusion and thresholding are notably understudied.
Feature fusion is normally done trivially and,
	when studied in depth,
	it rather focuses on solving the problems and computational challenges generated in previous phases of the algorithms.
For example, famous strategies such as edge tracking~\cite{Bergholm87,Papari07} or multiscale edge surfaces~\cite{Lindeberg98}
	are designed to cope with multiscale information,
	while Di Zenzo-Cumani operators~\cite{DiZenzo86,Cumani91} are presented because of the multidimensional nature of 
	colour inputs.
Some authors have indeed elaborated on the idea of intelligently fusing edge 
	cues in an interpretable manner~\cite{Law96},
	but they are rather exceptional.

In this work, we propose an edge detection algorithm which tries to simulate the 
	information flow in the human visual system for boundary detection.
This algorithm is faithful to the 4-phase schema presented above,
	and is guided by two principia:
	first, trying to keep all phases to the minimum complexity, in an attempt to mimic basic unitary behaviour in human neural tissue;
	second, using a sensible, adaptive feature fusion method to combine edge features. 	
The first principium renders into a simplistic, interpretable edge detection structure.
The second principium leads to edge cue fusion based on the generalizations of the Choquet integral~\cite{choquet1953},
	an aggregation operator that has been intensively studied in recent years~\cite{DIMURO202027}.

The application of the Choquet integral in Fuzzy Rule-Based Classification Systems~\cite{FRBCSs:Ishibuchi} led to an improvement of the system~\cite{LUCCA2018a}. This is due to the fact, that this aggregation when applied in the Fuzzy Reasoning Method (FRM)~\cite{axioms2020208} consider all the information related with the example and its relation with the fired rules. Observe for example, that the classical FRM of the Winning Rule (WR)~\cite{FRM:Cordon1998} just take into consideration one rule, the one having the largest relation with the example, and discard the remaining information.
	
In a similar way, we believe that the Choquet integral and its generalizations are a good computational representative of neuronal
	behaviour in terms of accumulating ionic charge, with the weights in the operator 
	representing the neuronal spiking profile.
As their properties have shown to succeed in a variety of tasks~\cite{DIMURO202027}, we think that in order to fill the gap between the HVS and edge detection the generalizations of Choquet integral permits to better detect the presence of edges as they consider the relation between the elements to be aggregated.
Hence, generalizations of the Choquet integral are taken as the keystone for the edge cue fusion.

In this context, the objectives of this work are:  
\begin{itemize}
    \item [(i)] Introduce a novel approach to the edge detection problem,  which consists in the application of the gravitational force smoothing, a technique that was  initially proposed by Marco-Detchart et al. in~\cite{Marco-Detchart2018}, combined with the use of generalizations of the Choquet integral in the feature blending phase of the Bezdek Breakdown Structure (BBS);
    \item [(ii)] Perform a quantitative evaluation of the obtained results, comparing them to different non-supervised edge detectors, like the Canny method \cite{Canny1986}, Fuzzy Morphology \cite{Gonzalez-Hidalgo2015} and Gravitational approaches \cite{Lopez-Molina2010b,Lopez-Molina2009}, using both the gravitational image smoothing mentioned in (i) and the         
        standard Gaussian \cite{nixon2019feature} image smoothing approach.
    \end{itemize}

This paper is organized as follows.
	In Section~\ref{sec:preliminaries}, we present  some preliminary and base concepts necessary to develop the paper. 
	In Section~\ref{Sec_Gen_choquet}, we present and discuss about some of the generalizations of the Choquet integral considered in the paper.
	Section~\ref{sec:proposal} is devoted to present the proposed methodology  for edge detection.
	In Section~\ref{sec:experiments}  we show some examples of edges obtained with our proposal, 
		along with the statistical  analysis and discussion of the obtained  results.
	Finally, in Section~\ref{sec:conclusions}, some conclusions are exposed along with future work.

\section{Preliminaries}\label{sec:preliminaries}

In this section, we recall some basic concepts about the theory of aggregation that will be used further on in the work.

In this paper, any function $F:[0,1]^n \to [0,1]$ is called a fusion function~\cite{Bustince2015}.  

\begin{defn}~\cite{LUCCA:CFintegrals} A bivariate fusion function $F:[0, 1]^2 \to [0, 1]$  with 0 as left annihilator element, that is, satisfying:
    \begin{description}
    	\item [(LAE)] $\forall y \in [0,1]: F(0,y) = 0$,
    \end{description} is said to be left 0-absorbent.
\end{defn}

Moreover, the following two basic properties are also important:

\begin{description}
\item [(RNE)] Right Neutral Element: $\forall x \in [0,1]: F(x,1) = x$;
\item [(LC)]  Left Conjunctive: $\forall x,y \in [0,1]: F(x,y)\leq x$.
\end{description}

Any bivariate  fusion function $F:[0,1]^2 \rightarrow[0,1]$ satisfying both \textbf{(LAE)} and \textbf{(RNE)} is called left 0-absorbent \textbf{(RNE)}-function.

A particular family of fusion functions is that of aggregation functions~\cite{Grabisch2009,Beliakov2016}.

\begin{defn}~\cite{Beliakov2007,Calvo2002} 
	A mapping $M:[0,1]^n\rightarrow [0,1]$ is an aggregation function if it is monotone non-decreasing in each of its components and satisfies
	the boundary conditions, $M(\mbox{\bf 0})=M(0,0,\ldots,0)=0$  and $M(\mbox{\bf 1})=M(1,1,\ldots,1)=1$.
\end{defn}

In the context of this work, the following two properties are important for aggregation functions:
\begin{description}
\item [(ID)] Idempotency:  $\forall (x, \ldots,x) \in [0,1]^n: M(x, \ldots,x) = x$;
\item [(AV)] Averaging behavior:  $\forall \mathbf{x}=(x_1, \ldots, x_n) \in [0,1]^n: \min \{x_1, \ldots, x_n\} \leq M(x_1, \ldots, x_n) \leq  \max \{x_1, \ldots, x_n\}$;
\end{description}

Observe that idempotency and averaging behavior are equivalent concepts in the context of aggregation functions.

\begin{defn}~\cite{Klement2000}
	An aggregation function $T:[0,1]^n\rightarrow [0,1]$ is said to be a t-norm if, for all $x,y,z \in [0,1]$, the following conditions hold:
\begin{enumerate}
		\item Commutativity: $T(x,y)=T(y,x)$;
		\item Associativity: $T(x,T(y,z))=T(T(x,y),z)$;
		\item Boundary conditions: $T(1,x) = T(x,1) = x$.
	\end{enumerate}
\end{defn}

\begin{defn}~\cite{Klement2000}
	An aggregation function $S:[0,1]^n\rightarrow [0,1]$ is said to be a t-conorm if, for all $x,y,z \in [0,1]$, the following conditions hold:
\begin{enumerate}
		\item Commutativity: $S(x,y)=S(y,x)$;
		\item Associativity: $S(x,T(y,z))=S(S(x,y),z)$;
		\item Boundary conditions: $S(x,0) = x$.
	\end{enumerate}
\end{defn}

A very relevant class of aggregation functions, which covers means \cite{Beliakov2016}, is that of Choquet integrals. To define these functions, we first need to recall the notion of fuzzy measure. In the following, consider  $N=\{1,2,\dots,n\}$.

\begin{defn}\label{def_measure}
	 A function $\mathfrak{m}: 2^{N} \rightarrow [0,1]$ is a fuzzy measure if, for all $X,Y \subseteq N$, it satisfies the following properties:
\begin{enumerate}
		\item { Increasingness}: if $X\subseteq Y$, then $\mathfrak{m}(X)\leq \mathfrak{m}(Y)$;
		\item Boundary conditions: $\mathfrak{m}(\emptyset) =0$ and $\mathfrak{m}(N) =1$.
	\end{enumerate}
\end{defn}

In the following, we present an example of fuzzy measure, namely, the power measure,  which is adopted in this work:
\begin{equation}\label{eq:cardExponencial}
	\mathfrak{m}_{q}(X) = \left ( \frac{|X|}{n} \right )^{q}, \ \mbox{with} \ q > 0,
\end{equation} where $|X|$ is the number of elements to be aggregated, $n$ the total number of elements and $q > 0$.  
We have selected this measure due to the fact that it is the one achieving the highest accuracy in classification problems~\cite{Lucca2016,LUCCA2018a}, where the exponent $q$ is learned by using a genetic algorithm. Here, however, we shall consider fixed values for $q$.

\begin{defn}\label{def_choquet}{\cite{Grabisch2009,Beliakov2016}} Let $\mathfrak{m}: 2^{N} \rightarrow [0,1]$ be a fuzzy measure. The discrete Choquet integral of  $\mathbf{x}=(x_1,\dots,x_n) \in [0,1]^n$ with respect to $\mathfrak{m}$ is defined as the function $C_{\mathfrak{m}} : [0,1]^n \rightarrow [0,1]$, given by
	\begin{equation*}
	C_{\mathfrak{m}} (\mathbf{x}) = \sum_{i=1}^{n} \left (x_{(i)} - x_{(i-1)} \right)\cdot  \mathfrak{m}\left(A_{(i)} \right),
	\label{eq:intChoquet}
	\end{equation*}
	where $\left(x_{(1)}, \ldots, x_{(n)}\right)$ is an increasing permutation on the input $\mathbf{x}$, that is,
	$ x_{(1)} \leq \ldots  \leq x_{(n)}$, with the convention that $x_{(0)} = 0$, and $A_{(i)} =  \{(i), \dots, (n)  \}$  is the subset of indices
	of the $n-i+1$ largest components of $\mathbf{x}$.
\end{defn}

The Choquet integral is idempotent and presents an averaging behavior. Observe that the Choquet integral is defined using a fuzzy measure, which allows it to take into consideration the relations or the interaction among the elements  to be aggregated (i.e., the components of an input $\mathbf{x}$). This is the key property of Choquet-like integrals.
Also remark that when the fuzzy measure only depends on the cardinality of the criteria the Choquet integral collapses to OWA operators~\cite{greco2011choquet}.

For some specific applications, imposing monotonicity might be too restrictive (e.g., the mode is not increasing with respect to all its arguments, but it is a valid function for certain applications). This consideration led Bustince~\etal~\cite{Bustince2015} to introduce the notion of directional monotonicity.

\begin{defn}~\cite{Bustince2015} \label{def:dirmon}
	Let $\vec{r}=(r_1,\dots,r_n)$ be a real $n$-dimensional vector, $\vec{r} \neq \vec{0}$. A fusion function $H:[0,1]^n \to [0,1]$ is $\vec{r}$-increasing if for all points $(x_1,\dots,x_n) \in [0,1]^n$ and for all $c>0$ such that $(x_1+cr_1,\dots,x_n+cr_n) \in [0,1]^n$ it holds
	\[
	H(x_1+cr_1,\dots,x_n+cr_n) \ge H(x_1,\dots,x_n)\; .
	\]
\end{defn}

That is, an $\vec{r}$-increasing function is a function which is increasing along the ray (direction) determined by the vector $\vec{r}$. For this reason, we say that $H$ is directionally monotone, or, more specifically, directionally $\vec{r}$-increasing. 
In fact in applications as classification tasks or decision making, the strict monotonicity is not necessary and this property has been proved to have no interference in the quality of the results~\cite{LUCCA2018,wieczynski2020generalizing}. This is why we intend to test this behaviour in our experiments.

\begin{defn}\label{def_preaggreg} (\cite{Lucca2016})
	A function $H:[0,1]^n \to [0,1]$ is said to be an n-ary pre-aggregation function  if the following conditions hold:
	\begin{description}
		\item[(1)]  $H$ is $\vec{r}$-increasing, for some $\vec{r} \in \mathbb{R}^n$, $\vec{r} \neq \vec{0}$;
		\item[(2)]  $H$ satisfies the boundary conditions: $H(0,\ldots, 0)=0$, $H(1,\ldots, 1)=1$.
	\end{description}
\end{defn}
	
Observe that idempotency and averaging behavior are not equivalent concepts in the context of pre-aggregation functions.

\section{Generalizations of the Choquet integral}\label{Sec_Gen_choquet}

The first generalization of the Choquet integral 
	was proposed by Lucca et al. in \cite{Lucca:TFS}. 
In that work, 
	the authors replace the product operator  
	of the Choquet integral by a t-norm $T$. 
This generalization 
	is referred to as $C_T$-integral, 
	and is defined as follows:

\begin{defn} \cite{Lucca2016} Let $\mathfrak{m}: 2^{N} \rightarrow [0,1]$ be a fuzzy measure and $T: [0,1]^2 \rightarrow [0,1]$ be a t-norm. Taking as basis the Choquet integral, we define the function $	\mathfrak{C}_{\mathfrak{m}}^{T}: [0,1]^n  \to [0,n]$, for all   $\mathbf{x} \in [0,1]^n$, by:
	\begin{equation}\label{eq:intChoquetPREAGG}
	\mathfrak{C}_{\mathfrak{m}}^T (\mathbf{x}) = \sum_{i=1}^{n}T\left(x_{(i)} - x_{(i-1)} ,  \mathfrak{m}\left(A_{(i)} \right) \right), \end{equation}
	where $\left(x_{(1)}, \ldots, x_{(n)}\right)$ is an increasing permutation on the input $\mathbf{x}$, that is,
	$ x_{(1)} \leq \ldots  \leq x_{(n)}$, with the convention that $x_{(0)} = 0$, and $A_{(i)} =  \{(i), \dots, (n)  \}$  is the subset of indices
	of the $n-i+1$ largest components of $\mathbf{x}$.
\end{defn}

\begin{theorem}\label{thm:gordo}\cite{Lucca2016}
Let $M:[0,1]^2 \to [0,1]$ be a function such that  for all $x,y \in [0,1]$ it
satisfies $M(x,y)\le x$, $M(x,1) = x$, $M(0,y) = 0$ and $M$ is (1,0)-increasing. Then, for any fuzzy measure $\mathfrak{m}$,  $C_{\mathfrak{m}}^M$ is a $(1,\ldots,1)$-pre-aggregation
function which is idempotent and averaging.
\end{theorem}

From the previous theorem,
	it follows that $C_T$-integrals are pre-aggregation functions,
	satisfying $(1,\ldots,1)$-increasingness, idempotency and averaging behaviour.

More recently, 
	Lucca et al.~\cite{LUCCA2018a}  generalize $C_T$-integrals 
	replacing the t-norm by a function $F:[0,1]^2 \rightarrow [0,1]$, 
	which, 
	under certain constraints,  
	enforce $C_T$-integrals to be pre-aggregation functions.  

\begin{defn}\cite{LUCCA2018a}\label{def_CFintegrals} 
	Let $F:[0,1]^2 \rightarrow [0,1]$ be a bivariate fusion function and $\mathfrak{m}: 2^{N} \rightarrow [0,1]$ be a fuzzy measure. The Choquet-like integral based on $F$  with respect to $\mathfrak{m}$, called $C_F$-integral,  is the function $\mathfrak{C}_{\mathfrak{m}}^{F} : [0,1]^n \rightarrow [0,1]$, defined, for all $\mathbf{x} \in [0,1]^n$,  by
	\begin{equation}\label{eq:CFintegrals}
		\mathfrak{C}_{\mathfrak{m}}^{F} (\mathbf{x}) = \min \left \{ 1, \sum_{i=1}^{n} F\left (  x_{(i)} -x_{(i-1)},  \mathfrak{m}\left(A_{(i)} \right) \right ) \right \}, 
	\end{equation}
	where $\left(x_{(1)}, \ldots, x_{(n)}\right)$ is an increasing permutation on the input $\mathbf{x}$, that is,
	$ x_{(1)} \leq \ldots  \leq x_{(n)}$, with the convention that $x_{(0)} = 0$, and $A_{(i)} =  \{(i), \dots, (n)  \}$  is the subset of indices
	of the $n-i+1$ largest components of $\mathbf{x}$.
\end{defn}

\begin{theorem}\label{teo-paf1}\cite{LUCCA2018a}  For any fuzzy measure $\mathfrak{m}: 2^{N} \rightarrow [0,1]$ and
left 0-absorbent \textbf{(RNE)}-function $F: [0,1]^2 \rightarrow [0,1]$, $\mathfrak{C}_{\mathfrak{m}}^{F} $ is a $(1, \ldots,1)$-pre-aggregation function.  Moreover, if $F$ also satisfies \textbf{(LC)}, then $\mathfrak{C}_{\mathfrak{m}}^{F}$ is idempotent and averaging. 
 \end{theorem}

\begin{theorem}\label{teo-paf2}\cite{LUCCA2018a}
For any fuzzy measure and left 0-absorbent $(1, 0)$-pre-aggregation function $F: [0, 1]^2\rightarrow [0, 1]$, $\mathfrak{C}_{\mathfrak{m}}^{F}$ is a-pre-aggregation function. Moreover, if $F$ also satisfies \textbf{(LC)}, then $\mathfrak{C}_{\mathfrak{m}}^{F}$ is idempotent and averaging. 
 \end{theorem}

Table~\ref{tab:tiposFunctions} presents some  of the best performer functions in our experimental study that can take up the role
	of the mapping $F$,
	combined with the corresponding fuzzy measure. 
Note that the generalizations are done 
	by replacing the \textit{x} and \textit{y} variables in each function by, 
	$(x_{(i)} - x_{(i-1)})$ and $\mathfrak{m}\left(A_{(i)}\right)$,
	respectively.

\begin{table*}
    \centering
    \begin{adjustbox}{max width=0.95\textwidth}
    \begin{tabular}{c c c}
    	\toprule
    	Choquet-like integral &  Base Function & Family\\
    	\midrule
    	$\mathfrak{C}_{\mathfrak{m}}^{C_F}$ & $C_F(x,y)=xy + x^2y(1-x)(1-y)$ (copula~\cite{afs2006})  & Averaging, Idempotent \\
    	
    	$\mathfrak{C}_{\mathfrak{m}}^{O_{B}}$ & $O_{B}(x,y) = \min \lbrace x\sqrt{y}, y\sqrt{x} \rbrace$ (overlap function~\cite{Bus10a,BDBB13})& Non-averaging, Non-Idempotent\\
    	
    	$\mathfrak{C}_{\mathfrak{m}}^{F_{BPC}}$ & $F_{BPC}(x,y) = xy^2$ (aggregation function)&  Averaging, Idempotent \\
    	
    	$\mathfrak{C}_{\mathfrak{m}}^{\text{Hamacher}}$ & $T_{HP}(x,y) =  \left\{ \begin{array}{ll}
			0 & \mbox{if} \ x = y = 0\\
			\frac{xy}{x+y-xy} & \mbox{otherwise}\\
			\end{array}
			\right.$ (t-norm)&  Averaging, Idempotent\\
\bottomrule
    \end{tabular}
	\end{adjustbox}
\caption{Families of $C_T$ and $C_F$-integrals used in this work, as generalizations of the Choquet integral.}
    \label{tab:tiposFunctions} 
\end{table*} 
\section{A methodology for  edge detection using $C_F$-integral-based feature fusion}\label{sec:proposal}

In our proposal, we intend to create an edge detection framework 
	based on (a) the computation of simplistic local edge features
	and (b) the aggregation of such features using generalizations of the Choquet integral.
In this manner, we intend to simulate the process of early detection of 
	visual primitives in the retina, followed by a neural combination of 
	such features in the early visual cortex~\cite{Marr82}.

We consider images to be functions $D: R \times C\mapsto L$, with $R= \{1,...,r\}$ and $C=\{1,...,c\}$, representing the set of rows and columns,
	and $L$ representing the set of tones of the image.
The set $L$ defines the type of image in question.
For binary images, $L=\{0,1\}$,
		whereas $L=\{0,...,255\}$ for gray-scale image pixels.
In the case of color images,
	$L$ is the Cartesian product of the tonal palettes at each of the color stimulus
	(e.g. $L=\{0,...,255\}^{3}$ for RGB images).
Assuming some given $D$,
	$\mathbb{I}_L$ represents the set of all images with a certain tonal palette $L$.
In this work, we consider real-valued grey-scale images, 
	\emph{i.e} images in  $\mathbb{I}_{[0,1]}$. 
	
Edge detection is often regarded as applying a kernel over the image to obtain a set of cues that represent the level of existence of an edge. 
In this sense, image features are obtained as a sliding window over each pixel of the image. Then those features are fused and filtered to obtain the final edge image.
As the features obtained are usually a set of values (in our case there are 8 features per pixel), we need to represent them with a unique value. This is why we fuse them using some type of operator, which in our proposal is done using the generalizations of the Choquet integral.

\begin{figure*}
	\centering
	\begin{tabular}{ccccc}
		{\small Original} & {\small Conditioning (P1)} & {\small Feature extraction (P2)} & {\small Blending (P3)} & {\small Scaling (P4)} \\
		
		\includegraphics[width=0.17\textwidth]{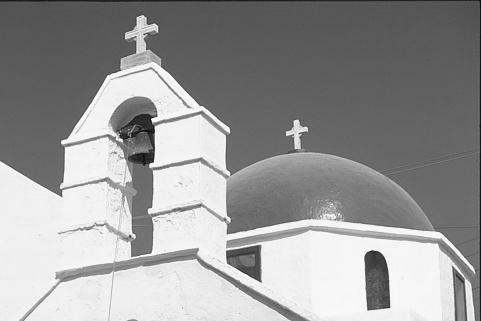} &
		\includegraphics[width=0.17\textwidth]{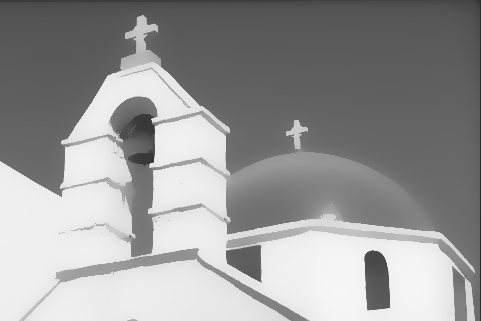} &
		\includegraphics[width=0.17\textwidth]{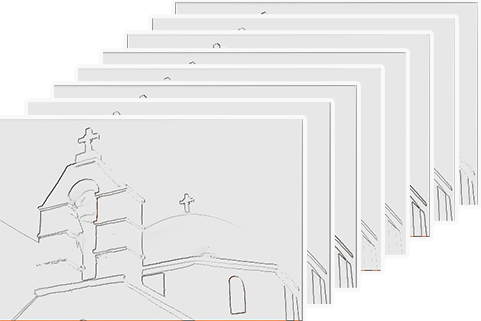} &
		\includegraphics[width=0.17\textwidth]{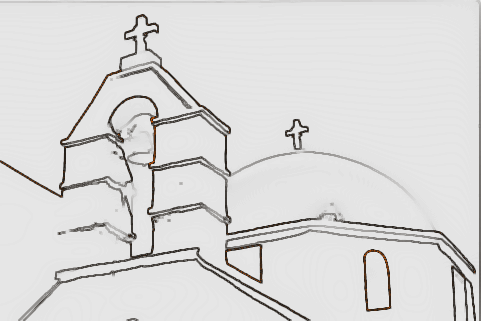}	&
		\fbox{\includegraphics[width=0.17\textwidth]{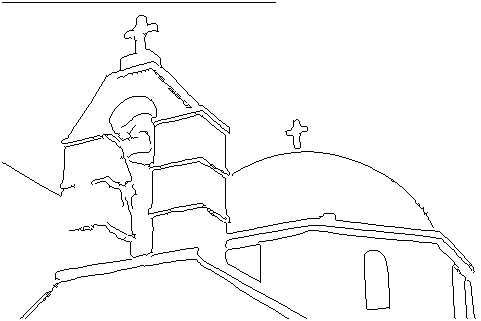}}	\\
	\end{tabular}
	\caption{Visual representation of the output of each of the phases in the BBS. Taking as an example the image 11835 from the BSDS, each phase has adhered to the proposal, as depicted in Section~\protect\ref{sec:proposal}.}
	\label{fig:Process}
\end{figure*}

Our edge detection framework 
	sticks to the Bezdek Breakdown Structure (BBS,~\cite{Bezdek98,LopezMolina12b}).
In this structure, the 
	process undergone by an image for edge detection consists of 
	four different phases: 
		image conditioning (P1), 
		feature extraction (P2), 
		blending (P3) and 
		scaling (P4).
This structure, 
	recap in Fig.~\ref{fig:Process},
	allows for the process of edge detection to be understood as four independent,
	coordinated phases.
In the \emph{conditioning} phase, the 
	image is enhanced for a better discrimination of its edges.
Examples of conditioning techniques include regularization and content-aware smoothing.
The \emph{feature extraction} phase consist of gathering local or semi-local cues 
	of the presence/absence of edges.
These cues are further fused at the \emph{blending} phase to produce scalar
	representations of the likeliness of an edge being present at each pixel.
Finally, the \emph{scaling} transforms the blended information into the desired
	representation, which is usually binary, 1-pixel wide edges.
Each of such phases is detailed individually in the upcoming sections.
		
\subsection{Conditioning (P1)}\label{subsec:conditioning}

As one of the steps in edge detection, image conditioning aims at removing noise or extra information not suitable for the edge extraction process. 
It is a crucial step as it helps to remove spurious artefacts in image acquisition, as well as to 
	reduce the effect of textures in edge detection.
Ideally, the conditioning process shall decrease the visual saliency of 
	non-edge structure, while preserving or maximizing the visibility of edges.
	
\begin{figure*}
	\centering
	\begin{tabular}{ccccc}
		Original & 
			$S_1$ & $S_2$ & $S_3$ & $S_4$ \\
		\includegraphics[width=0.17\textwidth]{imgs/118035} &
			\includegraphics[width=0.17\textwidth]{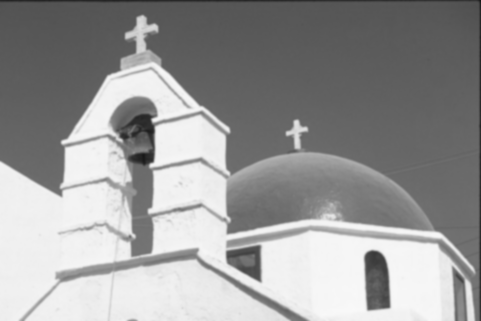} &
			\includegraphics[width=0.17\textwidth]{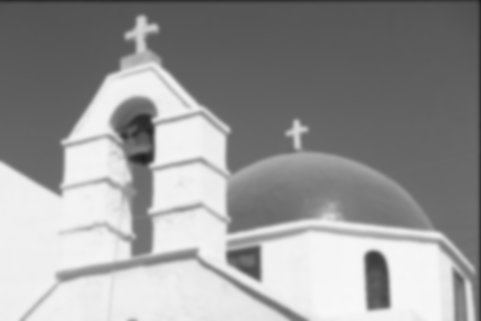} &
			\includegraphics[width=0.17\textwidth]{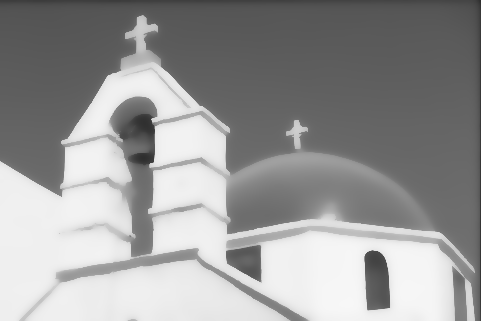} &
			\includegraphics[width=0.17\textwidth]{imgs/sm-118035-grav-it-50-0-0200-G-0-0500-cF-70-euc-euc}	
	\end{tabular}
	\caption{Original 118035 image from BSDS along with its result after image conditioning. The conditioning configurations are: Gaussian Smoothing with $\sigma=1$ ($S_1$) and $\sigma=2$ ($S_2$), and Gravitational Smoothing with the following configurations, $G=0.005, \omega_c = 20, t = 30$ ($S_3$) and $G=0.005, \omega_c = 70, t = 50$ ($S_4$).}
	\label{fig:imgSmoo}
\end{figure*}

We consider,
	in our proposal,
	two different conditioning methods. 
Firstly, the well-known Gaussian smoothing, 
	regularizes the image with a Gaussian pulse of standard deviation $\sigma$.
This $\sigma$ will control the level of smoothing over the image and generates a 2-D kernel which is applied over the image with a convolution.
The Gaussian smoothing is known to produce a blurring effect which affects edge regions.
 An example of this effect can be seen in the first two images of Figure~\ref{fig:imgSmoo}.
Secondly, 
	we use a more recent proposal for image regularization based on gravitational forces between particles 
		known as Gravitational Smoothing (\textit{GS})~\cite{Marco-Detchart2018}.
This approach,
	framed within the class of content-aware smoothing operators,
	iteratively simulates the grouping of pixels in a 5D, spatial-tonal space.
Similarly to the effects caused by Anisotropic Diffusion (in any of its proposed models~\cite{Perona90,Weickert98b}),
	Mean Shift~\cite{Comaniciu02} or Bilateral Filtering~\cite{Tomasi98},
	\textit{GS} succeeds in regularizing intra-object regions,
	while maintaining or improving the visibility of actual edges.
		
\begin{figure*}
	\centering
	\begin{tabular}{ccc}
		\begin{tikzpicture}
				\node[anchor=south west,inner sep=0] (image) at (0,0) {\includegraphics[width=0.33\textwidth]{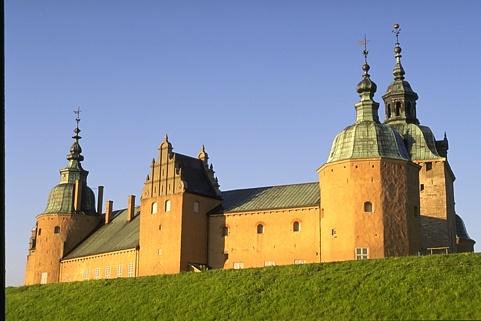}};
				\begin{scope}[x={(image.south east)},y={(image.north west)}]
				\draw[green,line width=0.15mm] (0.31,0.01) rectangle (0.33,0.99);
				\end{scope}
			\end{tikzpicture} &
		\includegraphics[width=0.28\textwidth]{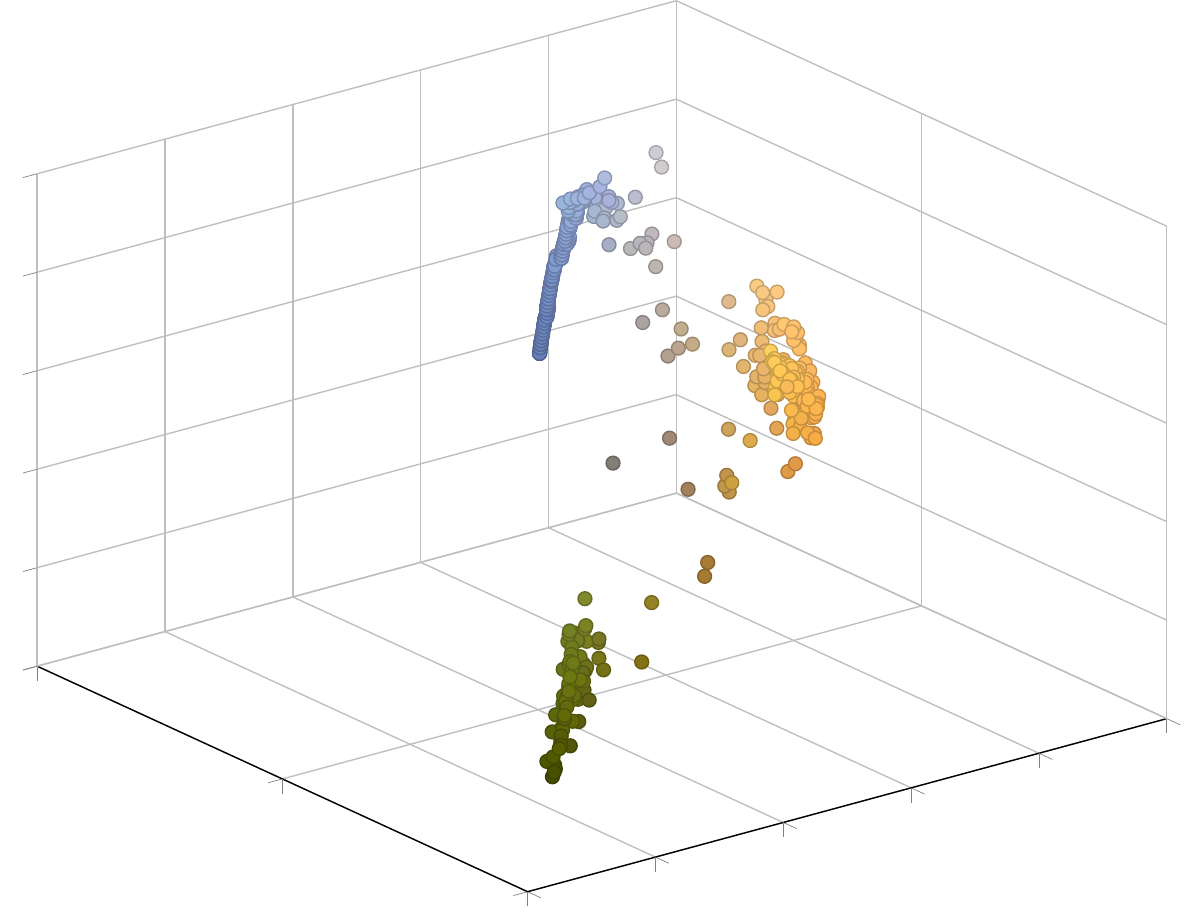} &
		\includegraphics[width=0.28\textwidth]{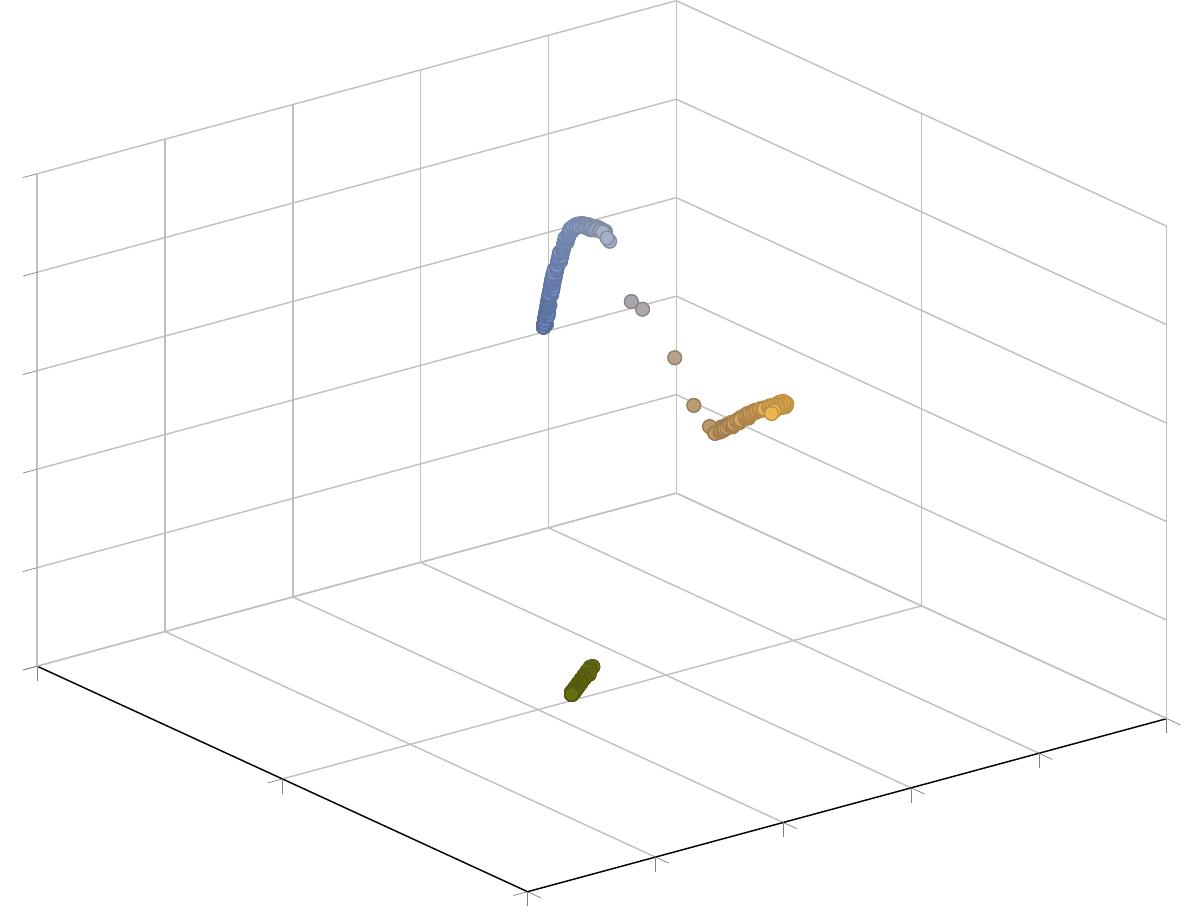}\\
	    \footnotesize{(a) Original image} &
	    \footnotesize{(b) Initial distribution of pixels} &
	    \footnotesize{(c) Final distribution of pixels}
	\end{tabular}
	\caption{Representation of the effect of Gravitational Smoothing (GS) in a color ~\cite{Marco-Detchart2018}. The figure displays (a) the original image, in which a vertical subregion is selected, (b) the initial distribution in the 3D RGB cubes of the pixels in that region and (c) the distribution of those same pixels after GS. It can be observed how tonal grouping is achieved, leading to regularized images.}
	\label{fig:scatter}
\end{figure*}

The regularization process (Figure~\ref{fig:scatter}) is done considering each pixel as a particle that exerts a force over every surrounding pixel.
The total force exerted over a particle is the sum of all the forces made by the remaining particles over it as indicated by Equation~\ref{eq:totforce}.

\begin{equation}\label{eq:totforce}
\lVert \vv{F_i} \rVert = \sum_{\substack{j \in N \\ i \neq j }} G \cdot \frac{m_{i} \cdot m_{j} }{\lVert \vv{r} \rVert ^{2}} \cdot \frac{\vv{r}}{\lVert \vv{r} \rVert}
\end{equation}

Where the mass of each particle is considered as $1$ and $\vv{r}$ is the 5-D distance in the spatial-tonal space.

Unlike Gaussian Smoothing, \textit{GS} 
	is highly configurable.	
According to~\cite{Marco-Detchart2018},
	we consider the following process-controlling parameters:
		$\omega_c$ controls the influence between the spatial and colour information of each pixel, 
		$t$ indicates the number of iterations to be performed,
		and $G$ is the gravitational constant. 
	The influence of these parameters on a conditioned image can be seen in the last two images in Figure~\ref{fig:imgSmoo}.

In this work, we consider the parameter settings in  Table~\ref{tab:configSmoo}.

\begin{table}
    \centering
    \begin{tabular}{c c c }
    	\toprule
    	Name & Smoothing method &  Parameter(s)\\
    	\midrule
    	$S_1$ & Gaussian & $\sigma = 1$\\
    	$S_2$ & Gaussian & $\sigma = 2$\\
    	$S_3$ & Gravitational & $G=0.05,~\omega_c = 20,~t = 30$\\
    	$S_4$ & Gravitational & $G=0.05,~\omega_c = 70,~t = 50$\\
    	\bottomrule
    \end{tabular}
    \caption{Configurations of the conditioning phase, as set for the experimental results in Section~\protect\ref{sec:results}.}
    \label{tab:configSmoo}
\end{table}

\subsection{Feature extraction (P2)}

At the feature extraction phase (Alg.~\ref{Alg:featExtraction}, Step~\ref{step:smoothing}),
	we intend to gather simplistic edge cues at each pixel.
The evident alternative would have been using Laplacian filters~\cite{Marr80,Marr82}, 
	hence mimicking the receptive fields at the retina.
However, we have opted out by an even simpler 
	set of edge cues.

At each pixel, 
	we select the 8-point immediate neighbourhood ($3\times3$ window) around each position $(x,y)\in D$ (\emph{i.e.} neighbours from $(x-1,y-1)$ to $(x+1,y+1)$). From such neighbourhood,
		we compute the absolute discrete difference (Alg.~\ref{Alg:featExtraction}-step~\ref{step:diff}) in the direction of each cardinal point, that is:
\begin{equation*}
	x_1=|a_{(x,y)}-a_{(x-1,y-1)}|,\ldots,x_8=|a_{(x,y)}-a_{(x+1,y+1)}|.
\end{equation*} 
	
This could have been done in more complex manners,
	e.g. anisotropic Gaussian kernels~\cite{Shui12,Wang19}.
However, the present configuration is maintained to preserve the simplicity of the process.

At each pixel,
	the 8 features are ordered increasingly (Alg.~\ref{Alg:featExtraction}-step~\ref{step:order}), so that
\begin{equation}
x_{\sigma_{(1)}} \leq x_{\sigma_{(2)}} \leq \ldots \leq  x_{\sigma_{(7)}} \leq x_{\sigma_{(8)}}.
\end{equation}
where $\sigma_{(1)}, \sigma_{(2)}, \ldots ,\sigma_{(8)}$ represents a permutation.

The output of the feature extraction is an image in $\mathbb{I}_{[0,1]^8}$.

\begin{algorithm}
\caption{Feature image extraction and blending using a Choquet integral generalization}
\label{Alg:featExtraction}
\begin{algorithmic}[1]
\renewcommand{\algorithmicrequire}{\textbf{Input:}}
\renewcommand{\algorithmicensure}{\textbf{Output:}}
\REQUIRE {A normalized grey-scale image $\mathbb{I}_g$ and a generalization of the Choquet integral}
\ENSURE  {A blended feature image $\mathbb{I}_f$.}
\STATE \label{step:smoothing} Smooth the image using the Gaussian and Gravitational Smoothing resulting in $\mathbb{I}_{sm}$;
\FOR {each pixel $(x,y)$ of $\mathbb{I}_{sm}$} 
\STATE \label{step:diff} Extract the corresponding features at given position by means of the absolute value of the difference between $\mathbb{I}_{sm}(x,y)$ and its $8$-neighbourhood ($3\times3$ window);
\STATE \label{step:order} Order the eight values of step \ref{step:diff}  in an increasing way;
\STATE \label{step:Choquet} Apply the generalization of the Choquet integral $\mathfrak{C}_{\mathfrak{m}}^{\mathbb{T}}$ or $\mathfrak{C}_{\mathfrak{m}}^{F}$ to the values obtained in step \ref{step:order} to fuse the different features;
\STATE Assign as intensity of the pixel $(x,y)$ of $\mathbb{I}_f$ the value obtained in step \ref{step:Choquet}.
\ENDFOR
\end{algorithmic}
\end{algorithm}

\subsection{Blending (P3)}

The blending phase 
	is in charge of aggregating, 
	at each pixel,
	the information gathered at the feature extraction phase into a scalar representative of the edges.
This phased aims at mimicking the 
		hierarchical composition of features in the early visual cortex (V1-V4),
		leading to the recognition of complex shapes from simple cues.
		
We understand that the proposed generalizations of the Choquet integral are adequate computational units 
	to simulate this process of neural aggregation.
Although,
	in general,
	any operator could be used (e.g. any aggregation operator~\cite{Beliakov2007,Perez20}),
	we believe that the generalizations of the Choquet integral are a better fit for the 
	non-linear spiking profile of neurons in the visual cortex.
In fact, their advantage relay in considering all the information related to the problem and modelling the relation between the cues being fused by means of the fuzzy measure.

In our method,
	edge features are blended (Alg.~\ref{Alg:featExtraction}-step~\ref{step:Choquet}) using
	either $\mathfrak{C}_{\mathfrak{m}}^{\mathbb{T}}$ or $\mathfrak{C}_{\mathfrak{m}}^{F}$.
The process is done at each pixel of the image obtaining a blended feature image $\mathbb{I}_{[0,1]}$. 
\subsection{Scaling (P4)}

In our proposal,
	we stick to the standard representation of edges as thin, binary lines. 
Hence, the scaling phase needs to convert the scalar, real-valued representation
	generated in the blending phase into a binary image.
This is done by applying non-maxima suppression~\cite{Rosenfeld70}, 
	considering cues orientation,
	in order to thin the peak values obtained from the blending step,
	and edge binarization using hysteresis.
Note that the process of setting the thresholding parameters
	needs to be carefully taken care of,
	and can be either determined using generic algorithms~\cite{Rosin01,Medina-Carnicer2011}
	or task-dependent procedures.

\section{Experimental framework}\label{sec:experiments}

In this section, we compare the 
	performance of our proposed framework with that
	by other well-established proposals in the literature.
First, in Section~\ref{quantification} we briefly introduce the dataset used for the experimentation along with the different quality measures used for the quantification of the results.
In Section~\ref{sec:competitors} we briefly describe the methods to whom we compare.
Then, in Section~\ref{sec:results} we present the quantitative 
	results in the comparison.

\subsection{Dataset and quantification of the results} \label{quantification}

\begin{figure*}
     \centering
         \begin{tabular}{ccccc}		
             \includegraphics[width=0.17\textwidth]{imgs/118035} & 
             \fbox{\includegraphics[width=0.17\textwidth]{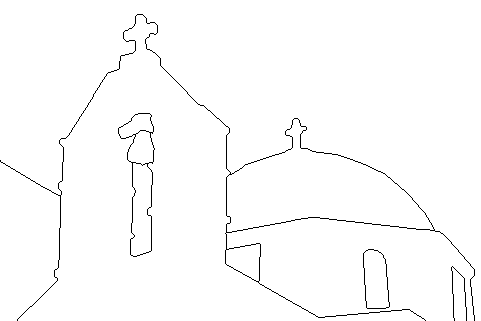}} & 
             \fbox{\includegraphics[width=0.17\textwidth]{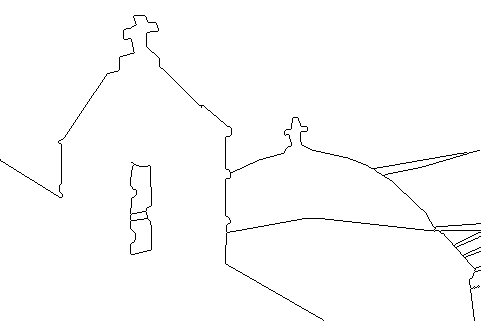}} &
             \fbox{\includegraphics[width=0.17\textwidth]{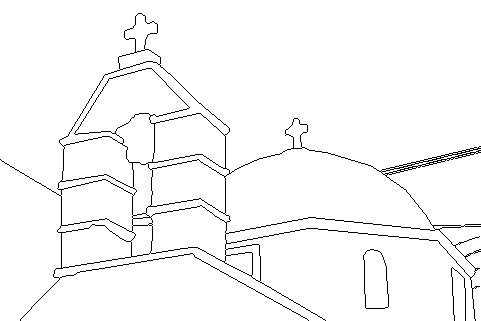}} &
             \fbox{\includegraphics[width=0.17\textwidth]{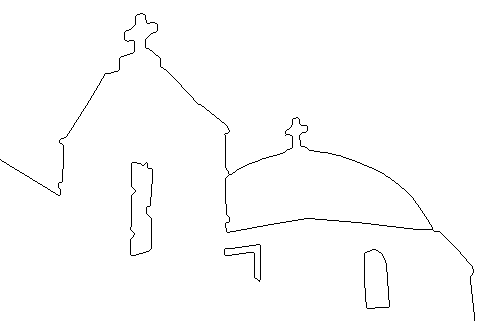}} 
         \end{tabular}
 \caption{Image 118035 from the BSDS dataset, together with four of its hand-labelled ground truth solutions.}
 \label{Fig:groundtruth}
 \end{figure*}

For our experiments, we have used the Test Set of the 
	Berkeley Segmentation Dataset and Benchmark (BSDS500,~\cite{Arbelaez2011}),
	which contains 200 natural images together with over 1000 ground truth labellings
	(see Fig.~\ref{Fig:groundtruth}).

The evaluation of an edge detection method is normally done 	
	by comparing its results with the hand-labelled results by human experts.
Considering both the output of an edge detection method and the
	ground truth labellings are represented as binary images, 
	edge image comparison can be taken as a binary classification problem.
This strategy renders into the generation of a binary classification matrix.
Note that,
	in order to compute the confusion matrix,
	an intelligent strategy needs to be taken in order to match the edges 
	in the automatically-generated results with those in the ground truth.
The reason is that edges which are at slightly displaced positions
	in different images should actually be accounted for as true positives,
	as long as such displacement is within a certain margin.
Among the different alternatives for computing the displacement-tolerant
	correspondence of edges,
	we have opted out by the standard procedure Estrada and Jepson~\cite{Estrada2009a},
	(the code is available at~\cite{KITT}). 
The spatial tolerance has been set to the 2.5\% of the length of the 
	image diagonal.

Normally, the confusion matrix in a binary image comparison procedure is 
	used as intermediate result for the generation of scalar descriptors
	of the similarity between the images.
We select the most-recognized descriptor, 
	which is the $\text{F}_{0.5}$ measure~\cite{Martin2004a,Lopez-Molina2013}.
This measure is computed from the precision and recall descriptors:
\begin{equation}
	\text{Prec}=\frac{\text{TP}}{\text{TP}+\text{FP}},~ 
	\text{Rec}=\frac{\text{TP}}{\text{TP}+\text{FN}}
\end{equation}

\noindent according to the following formula:
\begin{equation}
	\text{F}_\alpha=\frac{\text{Prec} \cdot \text{Rec}}{\alpha \cdot \text{Prec} + (1-\alpha) \cdot \text{Rec}} .
\end{equation}

The quantification of the result of an edge detection method at some given
	image is computed from the individual comparison to the images in its
	ground truth. Specifically, we keep the triplet $(\text{Prec},\text{Rec},\text{F}_{0.5})$ for
	the ground truth yielding the greatest $\text{F}_{0.5}$.
This process is repeated, then averaged, 
	for all images in the dataset.

\subsection{Methods in the experimental comparison} \label{sec:competitors}

The performance of our proposal is compared with that of other methods 
	in the literature, specifically:

\begin{itemize}
\item The Canny method~\cite{Canny1986}. The Canny method is normally implemented
	using a double filtering with Gaussian filters. First, at the conditioning phase,
	a zero-th order Gaussian pulse (standard deviation $\sigma_1$) is used to regularize the
	signal. Then, two orthogonal first order Gaussian filters (standard deviation $\sigma_2$)
	are used to estimate the partial derivatives of the underlying signal at each pixel.
	The blending is performed as the Euclidean norm of the gradient at each pixel.
	For this experiment, $\sigma_1 \in \{1,2\}$ and $\sigma_2= 2.25$, both being fairly
	standard values for images in the BSDS~\cite{Lopez-Molina2014,Medina-Carnicer2009}.

\item Gravitational Edge Detector based on a t-norm $T$~\cite{Lopez-Molina2010b,Lopez-Molina2009}. This method is based on the computation of the resulting local gravitational forces around each pixel. Those forces are expected to be small in magnitude in homogeneous regions, while being significantly larger
at boundaries. The method uses Gaussian smoothing for regularization,
	then computes the attractive gravitational forces by replacing the product in Newton's
	formulation by any t-norm or t-conorm.  In the present experiments, the functions used
	for such task are the probabilistic sum ($S_P(x,y)=x+y-xy$) and the maximum ($S_M(x,y)=\max(x,y)$), which are both t-conorms.

\item Fuzzy Morphology~\cite{Gonzalez-Hidalgo2015}. This method is based on a generalization of the morphological operators considering general t-conorms 
(t-norms) for erosion (dilation).
The morphological operators (erotion and dilation) consists in growing or shrinking an image. Thus, by enlarging or eroding away those areas where there could be a presence of edge new images with enlarged or shrinked cues are obtained.
Those pair of new images (\eg, substracting them) permit to obtain the final cues so that edges can be extracted.
The fuzzy erosion and dilation are based on the Schweizer-Sklar~\cite{Schweiser1961} t-norm and t-conorm, respectively,  which are defined,  for all $x,y \in [0,1]$, by: 

\begin{equation}
\begin{aligned}
	&T_{\lambda}^{\text{SS}}(x,y) =\left\lbrace
	\begin{array}{ll}
	\min(x,y), & \textup{if }  \lambda=-\infty,\\
	xy & \textup{if }  \lambda=0,\\
	T_{\text{D}}(x,y), & \textup{if }  \lambda=+\infty,\\
	(\max(x^\lambda+y^\lambda-1,0))^\frac{1}{\lambda}  & \textup{otherwise.}
	\end{array}
	\right.\\
	&\text{where }
	T_{\text{D}}(x,y)=\left\lbrace
	\begin{array}{ll}
	0, & \textup{if } x,y\in [0,1),  \\
	\min(x,y) & \textup{otherwise. } 
	\end{array}
	\right.
\end{aligned}
\label{Eq_SS}
\end{equation}

As advised in~\cite{Gonzalez-Hidalgo2015}, $\lambda=-5$ for this experiment.
We refer to this approach as $\text{FM}_\text{SS}$.

\end{itemize}

The process used for performance evaluation is as common as possible to all the methods compared. 
The conditioning is based on zero-th order Gaussian filters (as in the Canny method). Then, 
	after blending, the images are scaled using non-maxima suppression~\cite{Canny1986} and hysteresis~\cite{Medina-Carnicer2011}.

\subsection{Experimental results} \label{sec:results}

In this section we present the results obtained with our method compared to the ones obtained with well-known alternatives in the literature.
	We analyse both the qualitative and quantitative results,
		showing firstly a visual analysis where the different configurations are compared at a feature and boundary level and
		secondly a statistical analysis of the results as described in Section~\ref{quantification}.
	
\begin{figure*}
	\centering

	\begin{tabular}{cccccc}
		& $S_1$ & $S_2$ & $S_3$ & $S_4$ &\\
		
		\raisebox{0.9cm}{\rotatebox{90}{$\mathfrak{C}_{\mathfrak{m}}^{C_F}$}}&
		\includegraphics[width=0.2\textwidth]{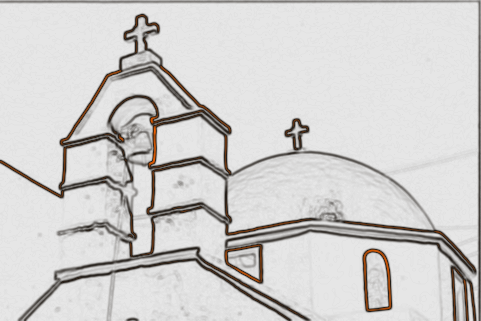} &
		\includegraphics[width=0.2\textwidth]{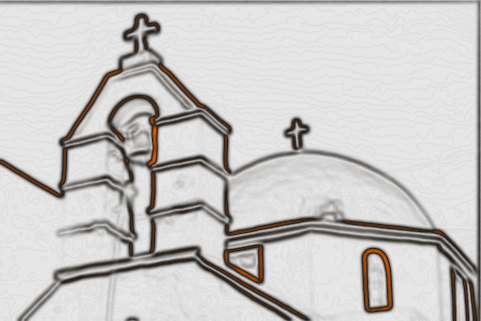} &
		\includegraphics[width=0.2\textwidth]{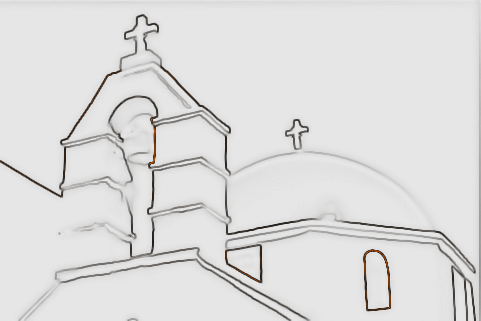} &
		\includegraphics[width=0.2\textwidth]{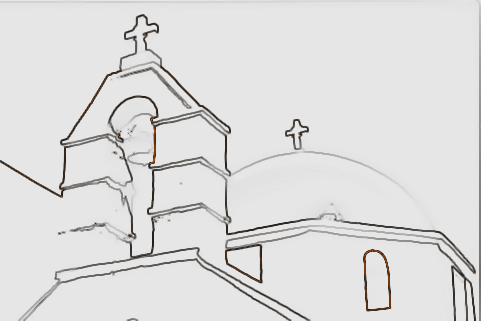} &
		\hspace{-0.35cm}\includegraphics[width=0.05\textwidth]{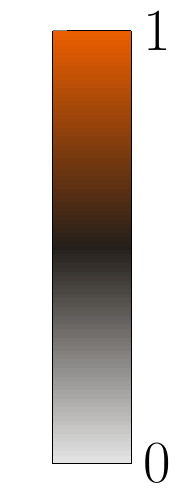}\\

		\raisebox{0.9cm}{\rotatebox{90}{$\mathfrak{C}_{\mathfrak{m}}^{O_{B}}$}}&
		\includegraphics[width=0.2\textwidth]{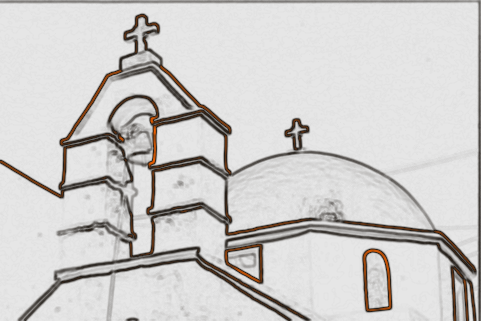} &
		\includegraphics[width=0.2\textwidth]{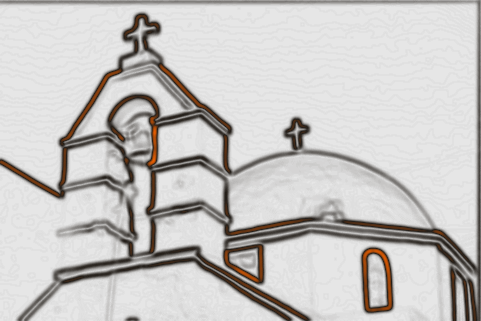} &
		\includegraphics[width=0.2\textwidth]{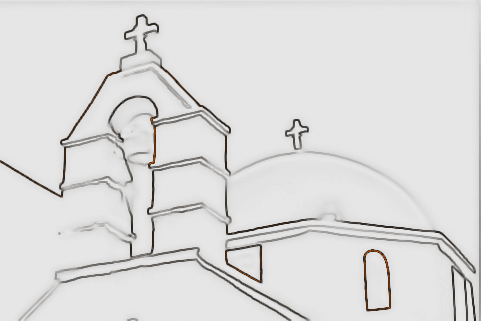} &
		\includegraphics[width=0.2\textwidth]{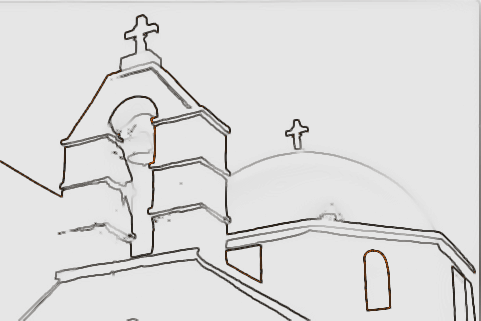} &\\
		
		\raisebox{0.5cm}{\rotatebox{90}{$\mathfrak{C}_{\mathfrak{m}}^{O_{FBPC}}$}}&
		\includegraphics[width=0.2\textwidth]{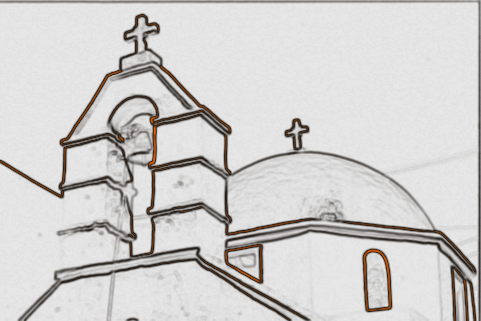} &
		\includegraphics[width=0.2\textwidth]{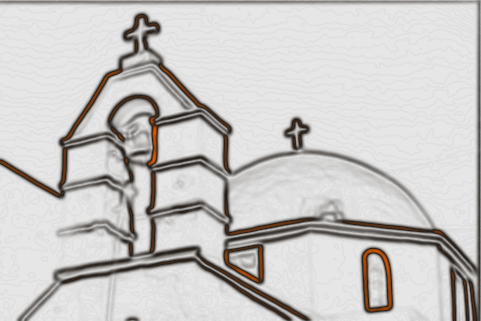} &
		\includegraphics[width=0.2\textwidth]{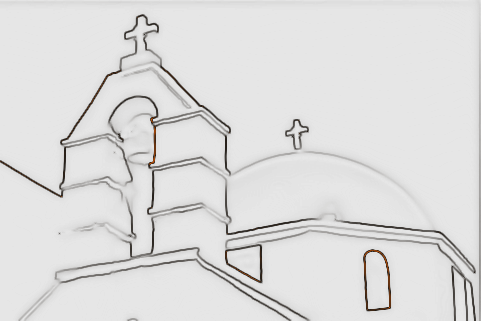} &
		\includegraphics[width=0.2\textwidth]{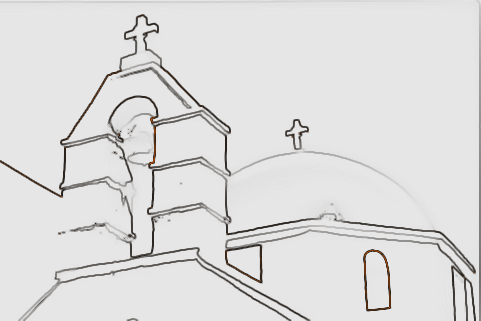} &\\
		
		\raisebox{0.6cm}{\rotatebox{90}{$\mathfrak{C}_{\mathfrak{m}}^{\text{Hamacher}}$}}&
		\includegraphics[width=0.2\textwidth]{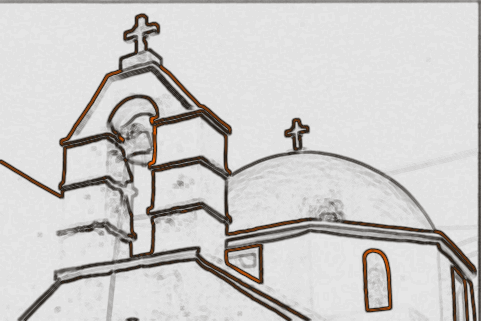} &
		\includegraphics[width=0.2\textwidth]{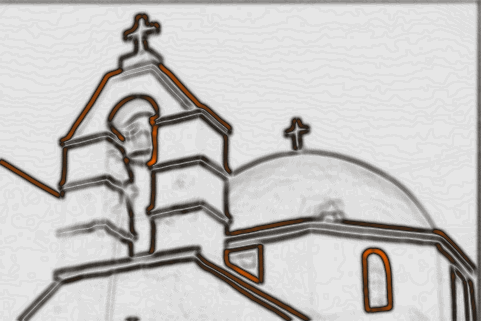} &
		\includegraphics[width=0.2\textwidth]{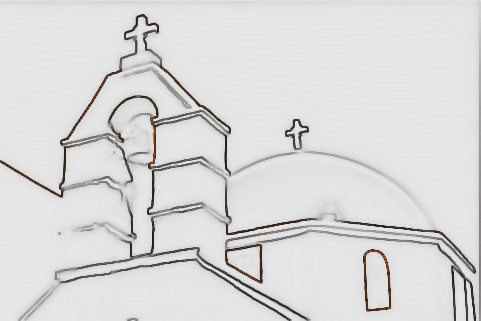} &
		\includegraphics[width=0.2\textwidth]{imgs/ft-118035-grav-it-50-0-0200-G-0-0500-cF-70-euc-euc-power-1-0000-CTM-F-hamacker} &\\
		
		\raisebox{0.6cm}{\rotatebox{90}{$\text{Canny}$}}&
		\includegraphics[width=0.2\textwidth]{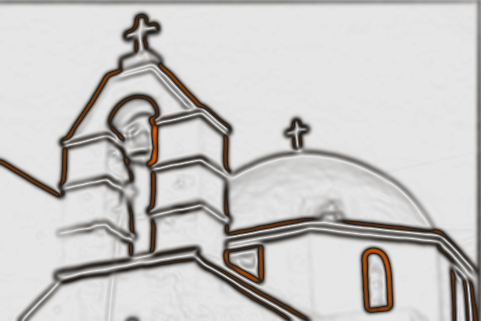} &
		\includegraphics[width=0.2\textwidth]{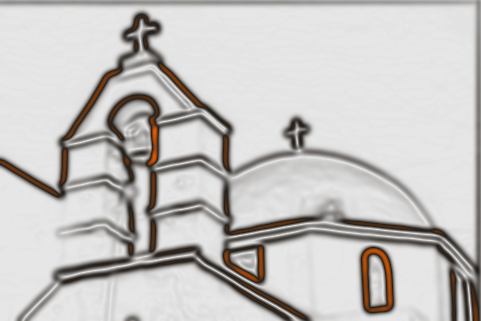} &
		\includegraphics[width=0.2\textwidth]{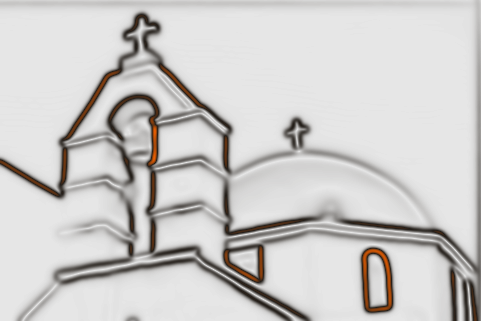} &
		\includegraphics[width=0.2\textwidth]{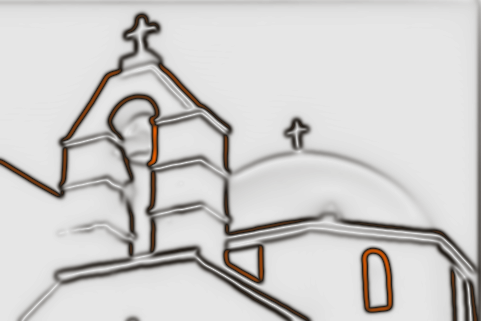} &\\
		
		\raisebox{0.7cm}{\rotatebox{90}{$\text{FM}_\text{SS}$}}&
		\includegraphics[width=0.2\textwidth]{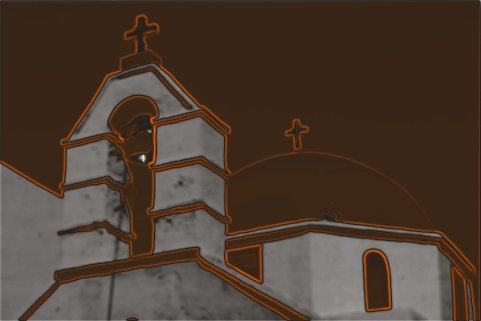} &
		\includegraphics[width=0.2\textwidth]{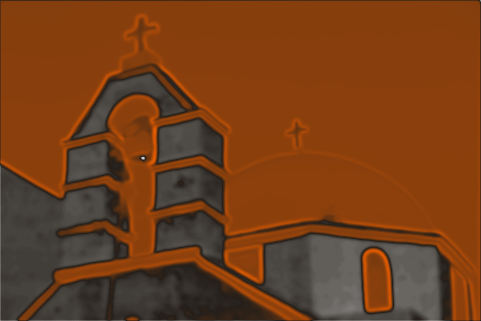} &
		\includegraphics[width=0.2\textwidth]{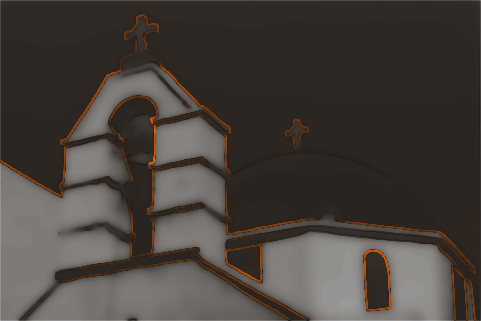} &
		\includegraphics[width=0.2\textwidth]{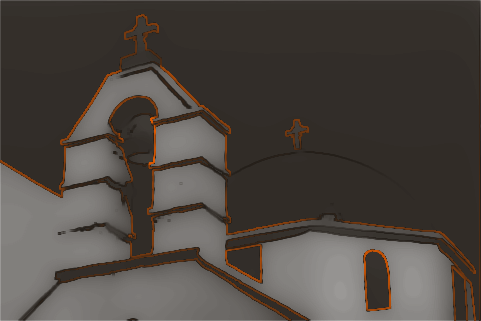} &\\
		
		\raisebox{0.7cm}{\rotatebox{90}{$G_{S_P}$}}&
		\includegraphics[width=0.2\textwidth]{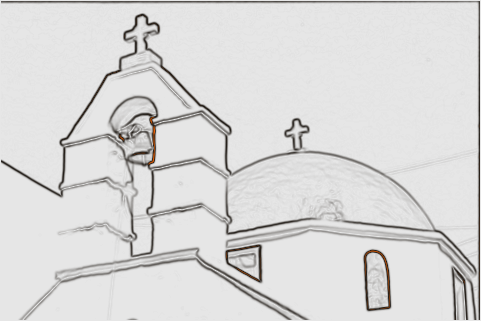} &
		\includegraphics[width=0.2\textwidth]{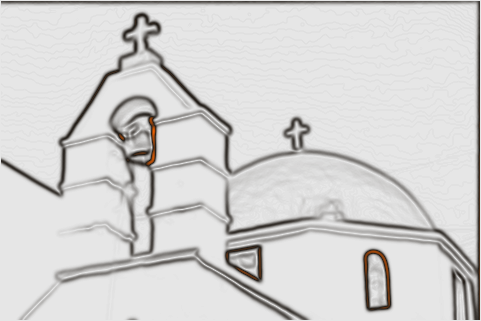} &
		\includegraphics[width=0.2\textwidth]{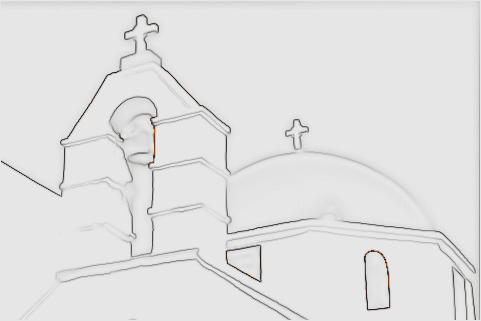} &
		\includegraphics[width=0.2\textwidth]{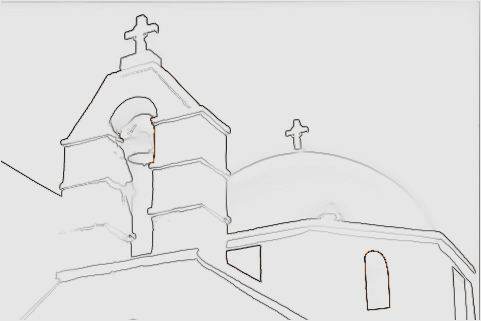} &\\
		
		\raisebox{0.7cm}{\rotatebox{90}{$G_{S_M}$}}&
		\includegraphics[width=0.2\textwidth]{imgs/ft-118035-gauss-1-0000-canny-2-2500} &
		\includegraphics[width=0.2\textwidth]{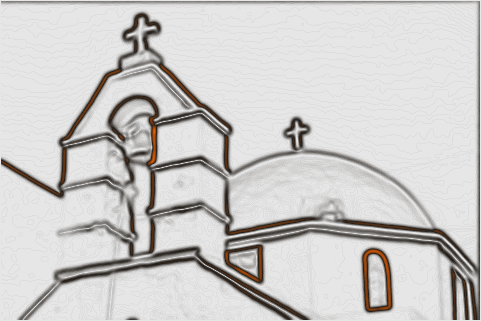} &
		\includegraphics[width=0.2\textwidth]{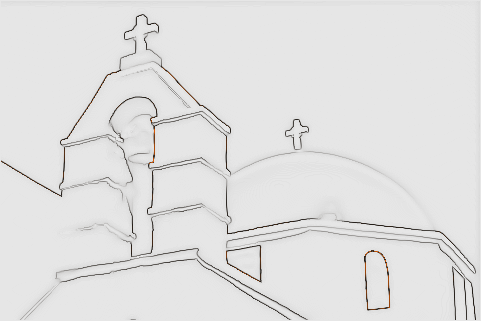} &
		\includegraphics[width=0.2\textwidth]{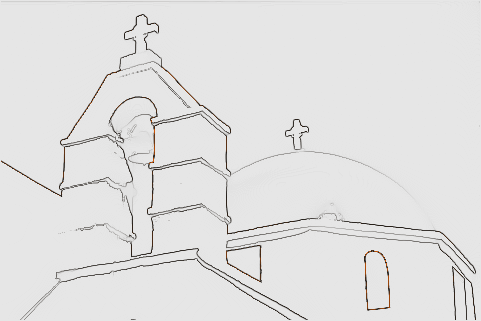} &
	\end{tabular}
\caption{Visual comparison of the different images obtained in the blending phase (P3) when applying the proposed functions in Table~\protect\ref{tab:tiposFunctions} along with methods from the literature as $Canny$, $FM_{SS}$, $G_{S_P}$, $G_{S_M}$}
	\label{fig:imgFeat}
\end{figure*} 
\begin{figure*}
	\centering
	\begin{tabular}{ccccccccc}
	   	& $S_1$ & $S_2$ & $S_3$ & $S_4$\\
 
		\raisebox{0.9cm}{\rotatebox{90}{$\mathfrak{C}_{\mathfrak{m}}^{C_F}$}}&
 		\fbox{\includegraphics[width=0.2\textwidth]{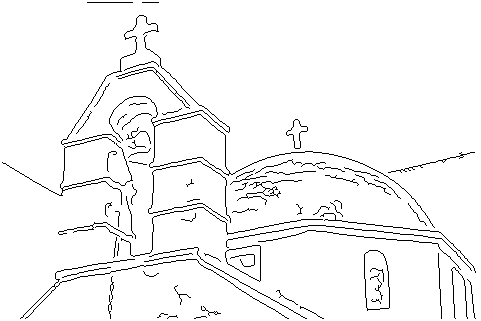}} &
 		\fbox{\includegraphics[width=0.2\textwidth]{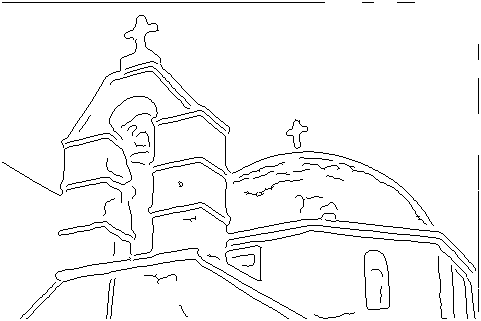}} &
 		\fbox{\includegraphics[width=0.2\textwidth]{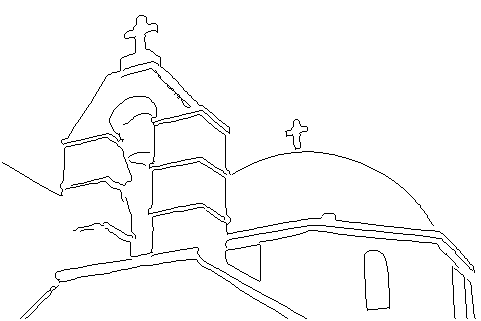}} &
 		\fbox{\includegraphics[width=0.2\textwidth]{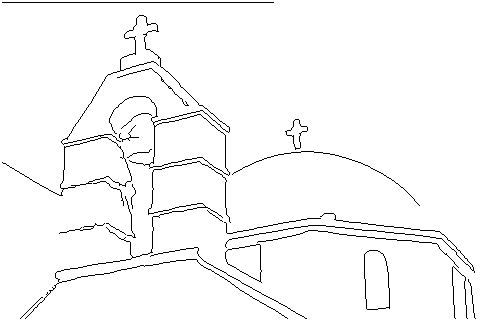}} \\

		\raisebox{0.9cm}{\rotatebox{90}{$\mathfrak{C}_{\mathfrak{m}}^{O_{B}}$}}&
		\fbox{\includegraphics[width=0.2\textwidth]{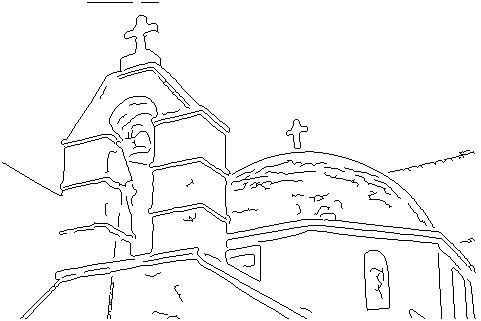}} &
 		\fbox{\includegraphics[width=0.2\textwidth]{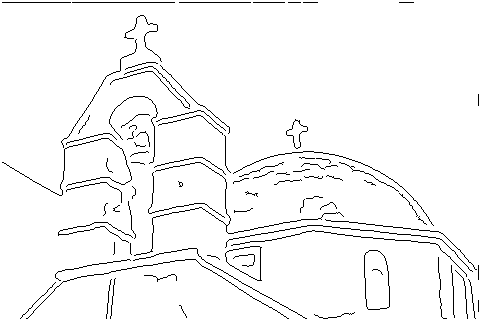}} &
 		\fbox{\includegraphics[width=0.2\textwidth]{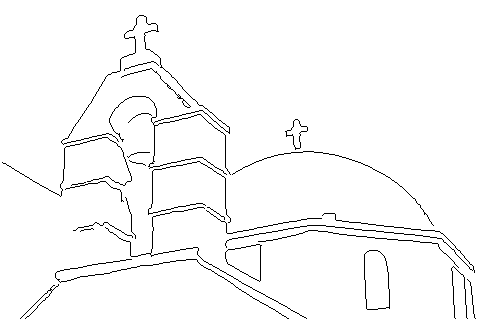}} &
 		\fbox{\includegraphics[width=0.2\textwidth]{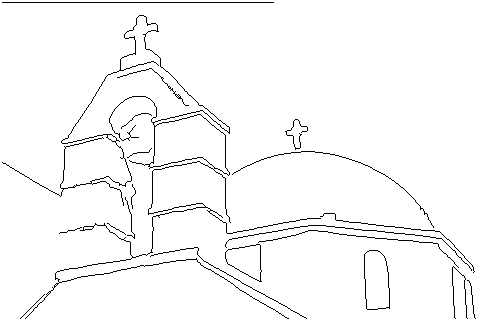}} \\
		
		\raisebox{0.5cm}{\rotatebox{90}{$\mathfrak{C}_{\mathfrak{m}}^{O_{FBPC}}$}}&
		\fbox{\includegraphics[width=0.2\textwidth]{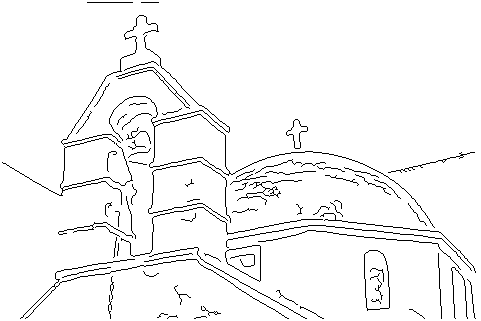}} &
 		\fbox{\includegraphics[width=0.2\textwidth]{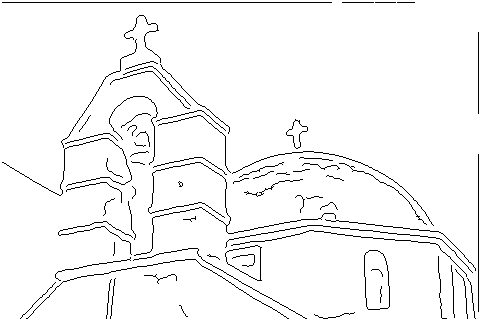}} &
 		\fbox{\includegraphics[width=0.2\textwidth]{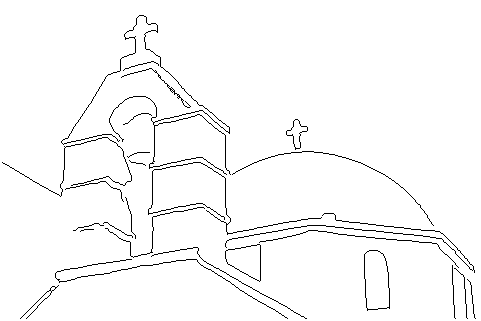}} &
 		\fbox{\includegraphics[width=0.2\textwidth]{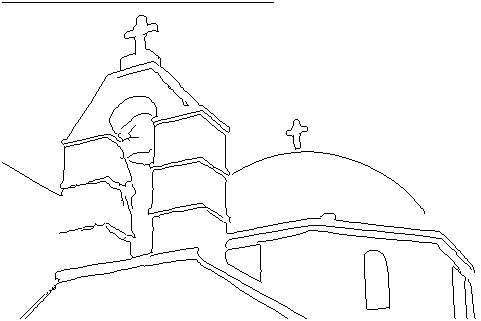}} \\
		
		\raisebox{0.6cm}{\rotatebox{90}{$\mathfrak{C}_{\mathfrak{m}}^{\text{Hamacher}}$}}&
		\fbox{\includegraphics[width=0.2\textwidth]{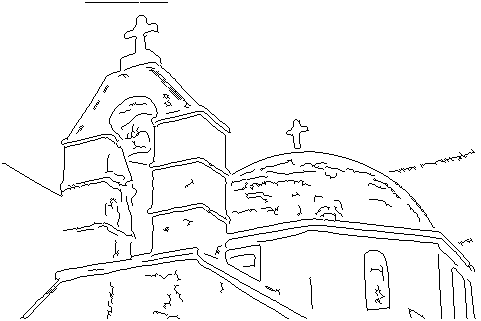}} &
 		\fbox{\includegraphics[width=0.2\textwidth]{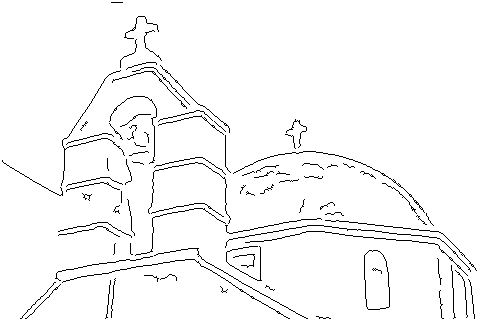}} &
 		\fbox{\includegraphics[width=0.2\textwidth]{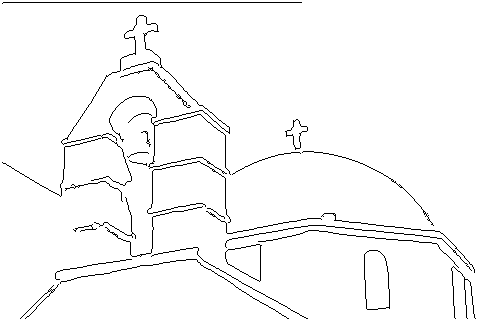}} &
 		\fbox{\includegraphics[width=0.2\textwidth]{imgs/bdry-118035-grav-it-50-0-0200-G-0-0500-cF-70-euc-euc-power-1-0000-CTM-F-hamacker}} \\
		
		\raisebox{0.6cm}{\rotatebox{90}{$\text{Canny}$}}&
		\fbox{\includegraphics[width=0.2\textwidth]{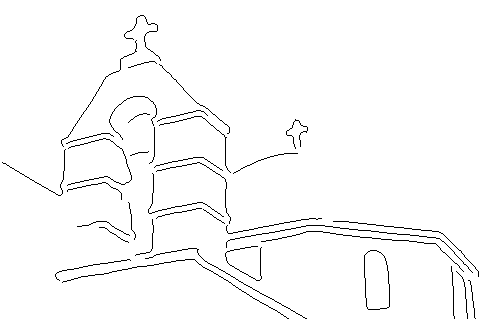}} &
 		\fbox{\includegraphics[width=0.2\textwidth]{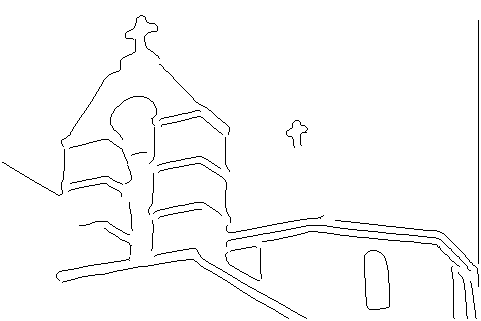}} &
 		\fbox{\includegraphics[width=0.2\textwidth]{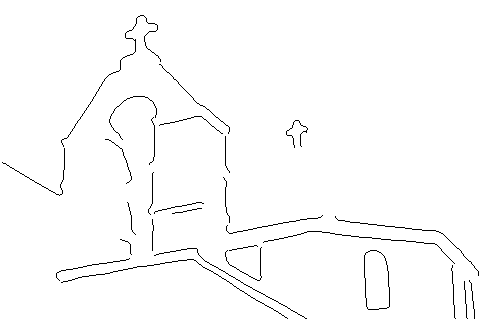}} &
 		\fbox{\includegraphics[width=0.2\textwidth]{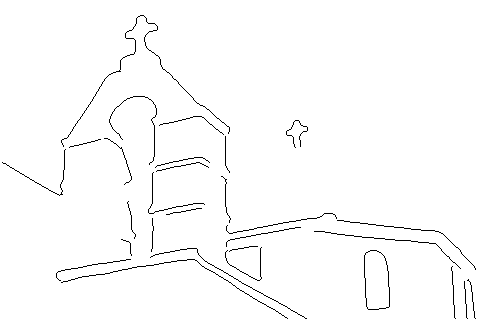}} \\
		
		\raisebox{0.7cm}{\rotatebox{90}{$\text{FM}_\text{SS}$}}&
		\fbox{\includegraphics[width=0.2\textwidth]{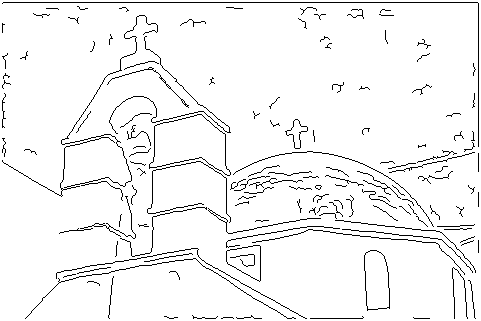}} &
 		\fbox{\includegraphics[width=0.2\textwidth]{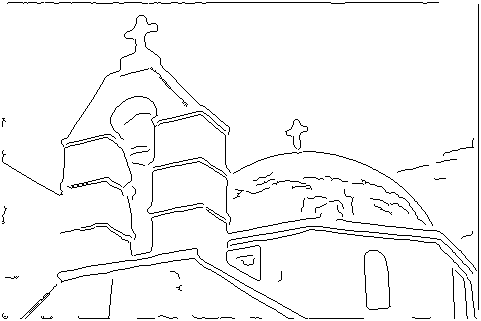}} &
 		\fbox{\includegraphics[width=0.2\textwidth]{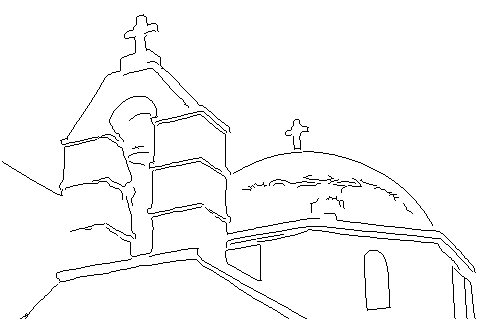}} &
 		\fbox{\includegraphics[width=0.2\textwidth]{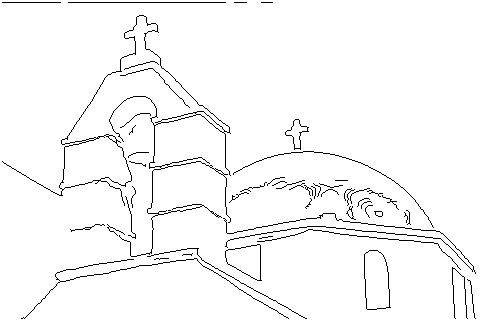}} \\
				
		\raisebox{0.7cm}{\rotatebox{90}{$G_{S_P}$}}&
		\fbox{\includegraphics[width=0.2\textwidth]{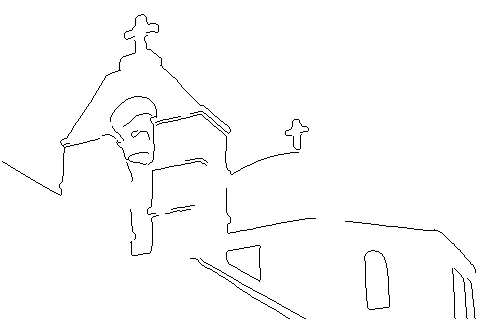}} &
 		\fbox{\includegraphics[width=0.2\textwidth]{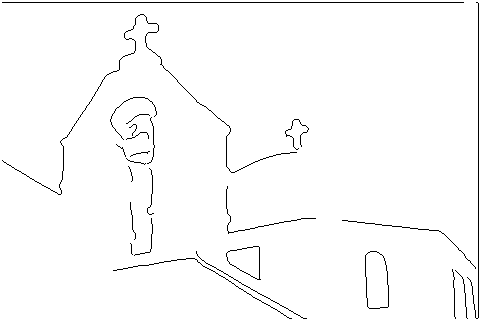}} &
 		\fbox{\includegraphics[width=0.2\textwidth]{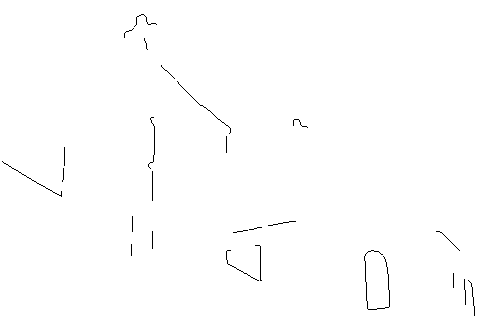}} &
 		\fbox{\includegraphics[width=0.2\textwidth]{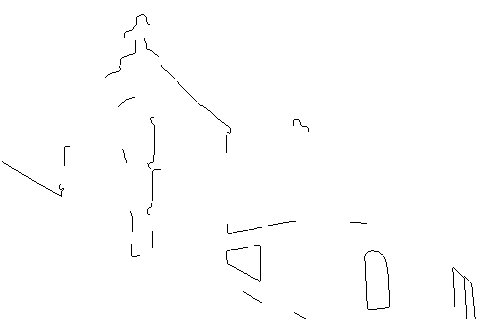}} \\
		
		\raisebox{0.7cm}{\rotatebox{90}{$G_{S_M}$}}&
		\fbox{\includegraphics[width=0.2\textwidth]{imgs/bdry-118035-gauss-1-0000-canny-2-2500}} &
 		\fbox{\includegraphics[width=0.2\textwidth]{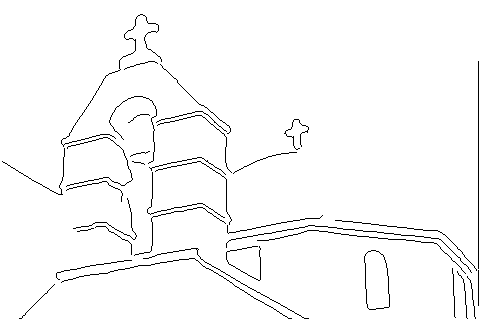}} &
 		\fbox{\includegraphics[width=0.2\textwidth]{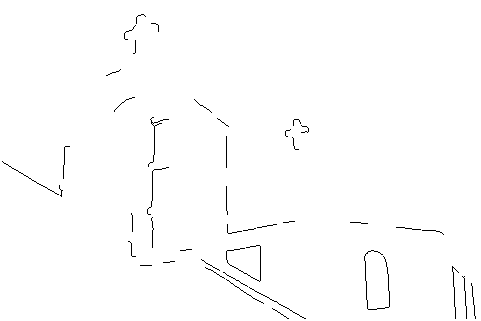}} &
 		\fbox{\includegraphics[width=0.2\textwidth]{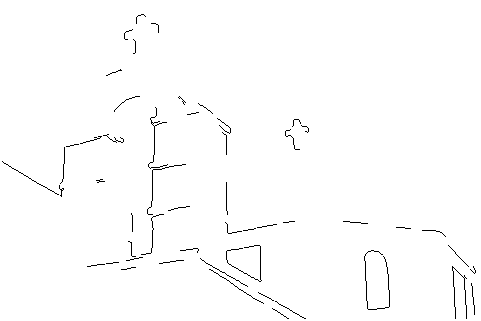}} \\	
	\end{tabular}
\caption{Visual comparison of the different boundary images obtained when applying the proposed functions in Table~\protect\ref{tab:tiposFunctions} along with methods from the literature as $\text{Canny}$, $\text{FM}_\text{SS}$, $G_{S_P}$, $G_{S_M}$}
	\label{fig:imgBdry}
\end{figure*} 
\begin{figure*}
	\centering
\begin{tabular}{ccccc}
	   	\raisebox{0.6cm}{\rotatebox{90}{Original}}&
 		\includegraphics[width=0.2\textwidth]{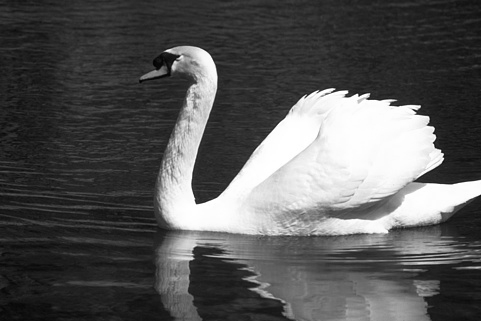} &
 		\includegraphics[width=0.2\textwidth]{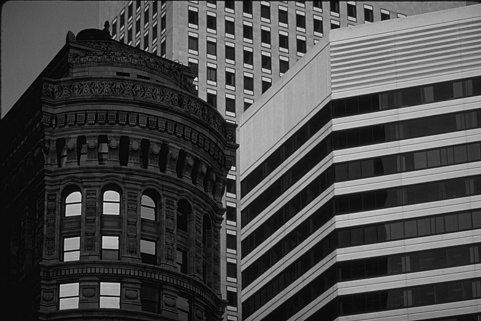} &
 		\includegraphics[width=0.2\textwidth]{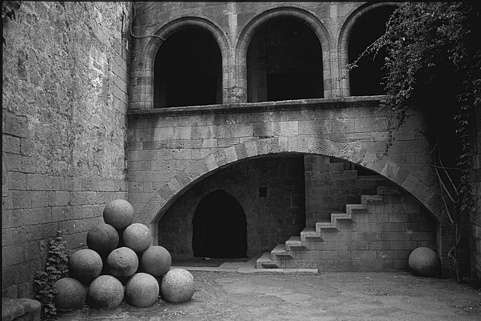} &
 		\includegraphics[width=0.2\textwidth]{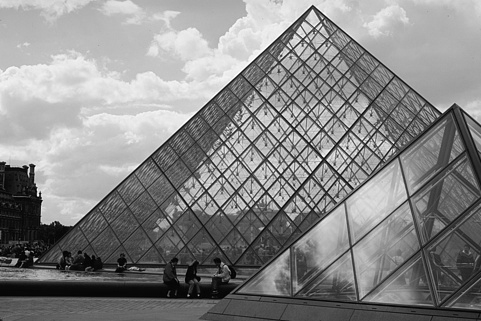} \\
 
		\raisebox{0.9cm}{\rotatebox{90}{$\mathfrak{C}_{\mathfrak{m}}^{C_F}$}}&
 		\fbox{\includegraphics[width=0.2\textwidth]{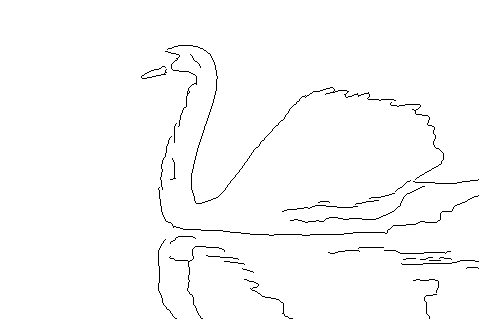}} &
 		\fbox{\includegraphics[width=0.2\textwidth]{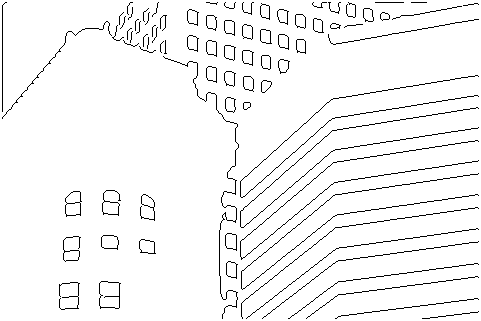}} &
 		\fbox{\includegraphics[width=0.2\textwidth]{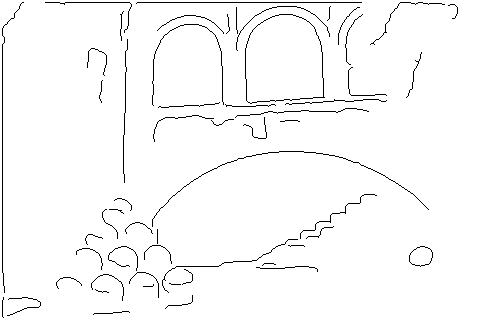}} &
 		\fbox{\includegraphics[width=0.2\textwidth]{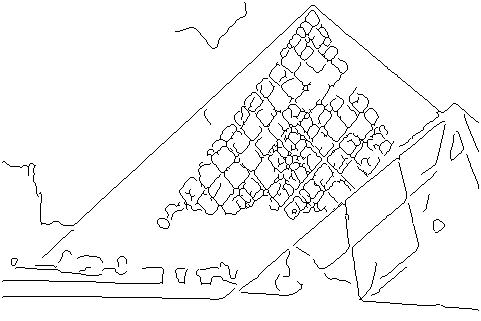}} \\

		\raisebox{0.9cm}{\rotatebox{90}{$\mathfrak{C}_{\mathfrak{m}}^{O_{B}}$}}&
		\fbox{\includegraphics[width=0.2\textwidth]{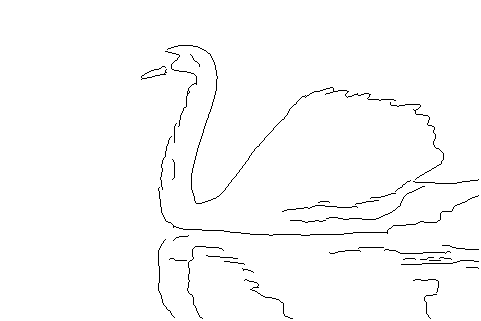}} &
 		\fbox{\includegraphics[width=0.2\textwidth]{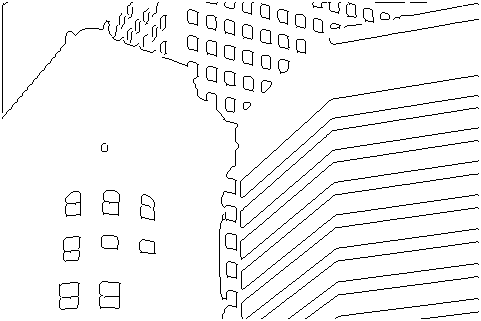}} &
 		\fbox{\includegraphics[width=0.2\textwidth]{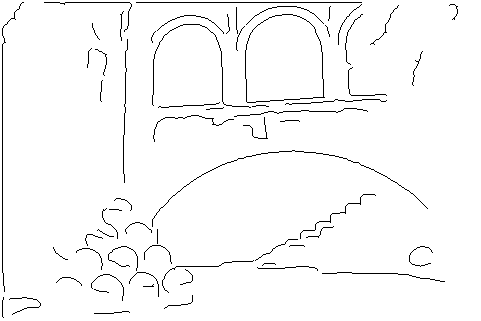}} &
 		\fbox{\includegraphics[width=0.2\textwidth]{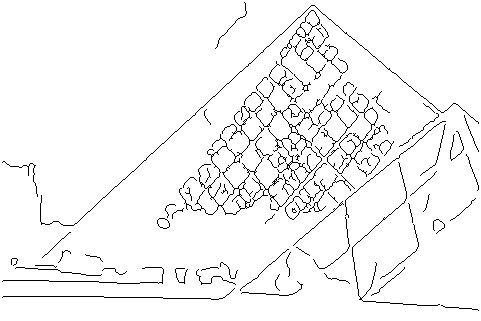}} \\
		
		\raisebox{0.5cm}{\rotatebox{90}{$\mathfrak{C}_{\mathfrak{m}}^{O_{FBPC}}$}}&
		\fbox{\includegraphics[width=0.2\textwidth]{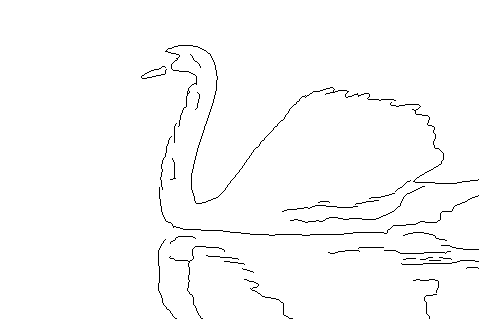}} &
 		\fbox{\includegraphics[width=0.2\textwidth]{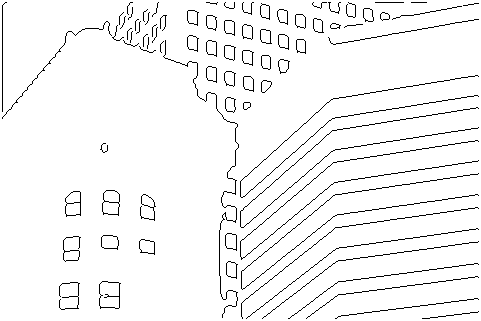}} &
 		\fbox{\includegraphics[width=0.2\textwidth]{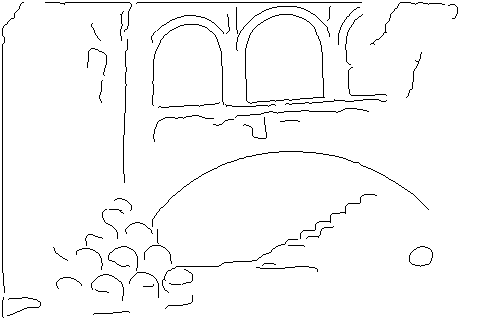}} &
 		\fbox{\includegraphics[width=0.2\textwidth]{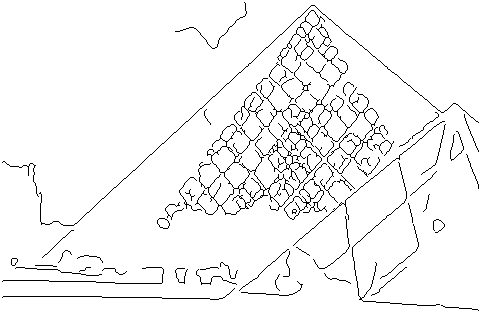}} \\
		
		\raisebox{0.6cm}{\rotatebox{90}{$\mathfrak{C}_{\mathfrak{m}}^{\text{Hamacher}}$}}&
		\fbox{\includegraphics[width=0.2\textwidth]{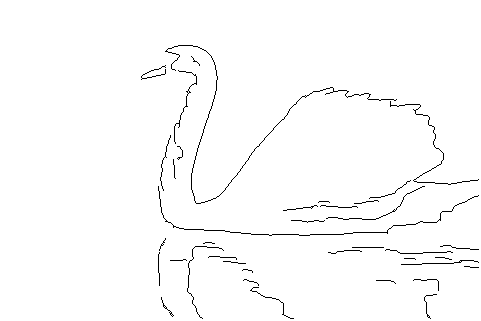}} &
 		\fbox{\includegraphics[width=0.2\textwidth]{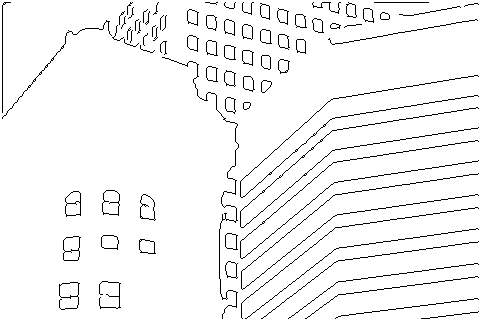}} &
 		\fbox{\includegraphics[width=0.2\textwidth]{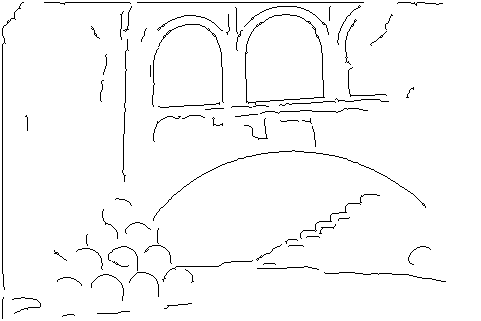}} &
 		\fbox{\includegraphics[width=0.2\textwidth]{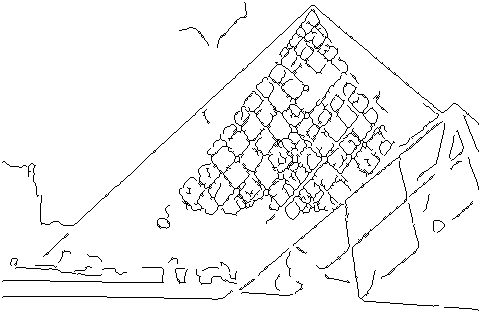}} \\
		
		\raisebox{0.6cm}{\rotatebox{90}{$\text{Canny}$}}&
		\fbox{\includegraphics[width=0.2\textwidth]{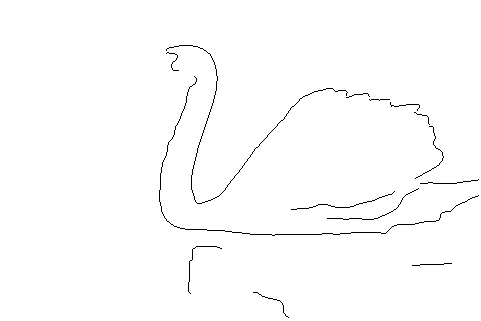}} &
 		\fbox{\includegraphics[width=0.2\textwidth]{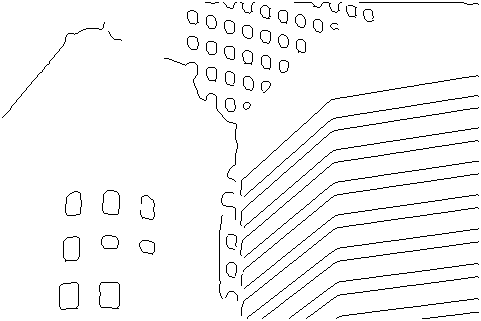}} &
 		\fbox{\includegraphics[width=0.2\textwidth]{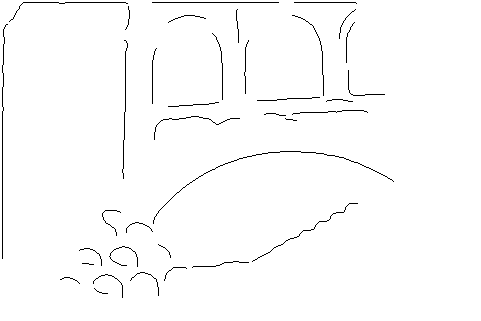}} &
 		\fbox{\includegraphics[width=0.2\textwidth]{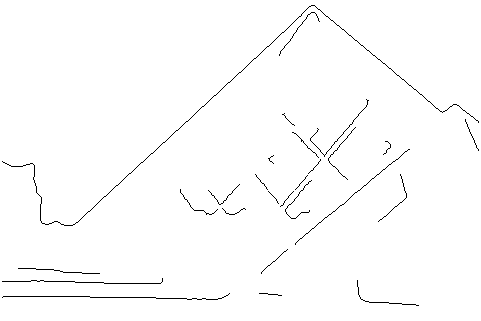}} \\
		
		\raisebox{0.7cm}{\rotatebox{90}{$\text{FM}_\text{SS}$}}&
		\fbox{\includegraphics[width=0.2\textwidth]{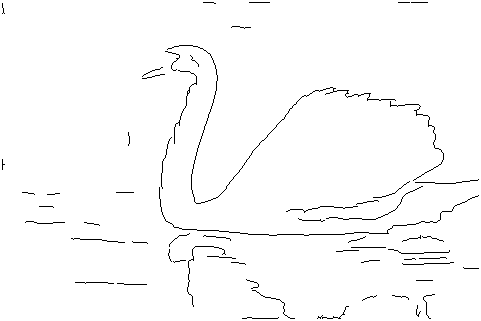}} &
 		\fbox{\includegraphics[width=0.2\textwidth]{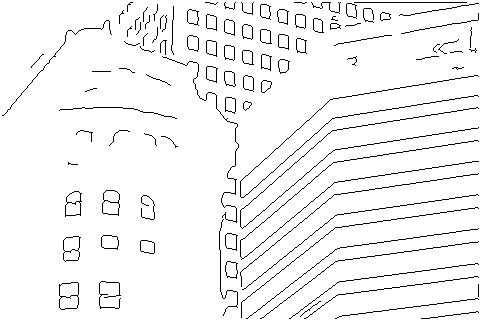}} &
 		\fbox{\includegraphics[width=0.2\textwidth]{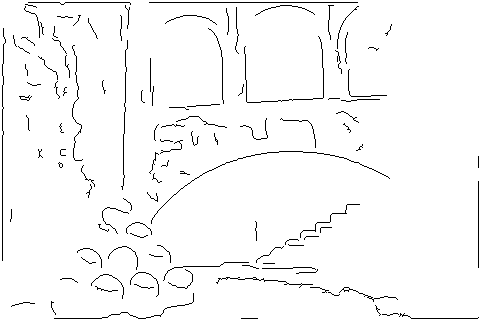}} &
 		\fbox{\includegraphics[width=0.2\textwidth]{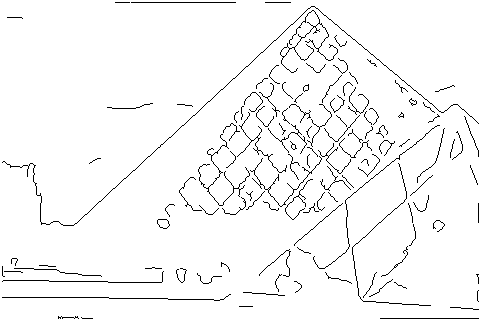}} \\
				
		\raisebox{0.7cm}{\rotatebox{90}{$G_{S_P}$}}&
		\fbox{\includegraphics[width=0.2\textwidth]{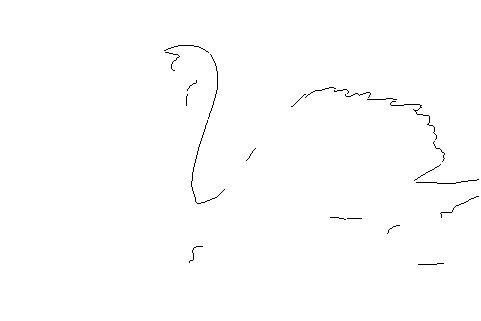}} &
 		\fbox{\includegraphics[width=0.2\textwidth]{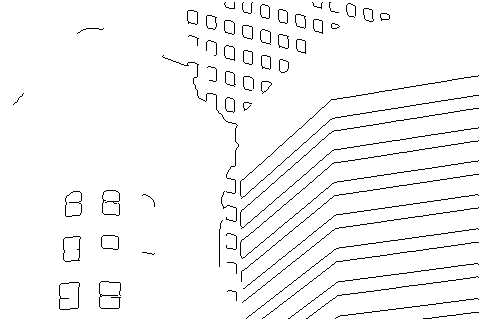}} &
 		\fbox{\includegraphics[width=0.2\textwidth]{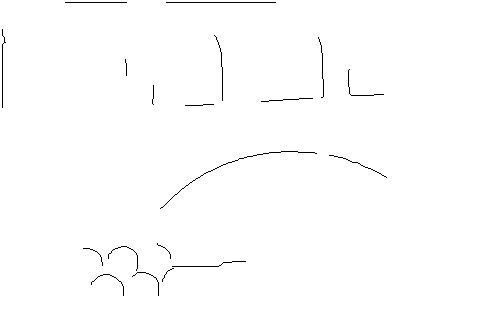}} &
 		\fbox{\includegraphics[width=0.2\textwidth]{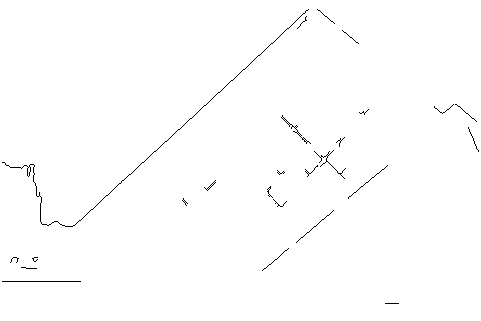}} \\
		
		\raisebox{0.7cm}{\rotatebox{90}{$G_{S_M}$}}&
		\fbox{\includegraphics[width=0.2\textwidth]{imgs/extra/bdry-8068-grav-it-30-0-0500-G-0-0500-cF-20-euc-euc-canny-2-2500}} &
 		\fbox{\includegraphics[width=0.2\textwidth]{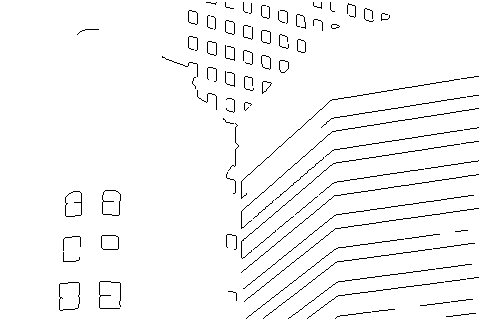}} &
 		\fbox{\includegraphics[width=0.2\textwidth]{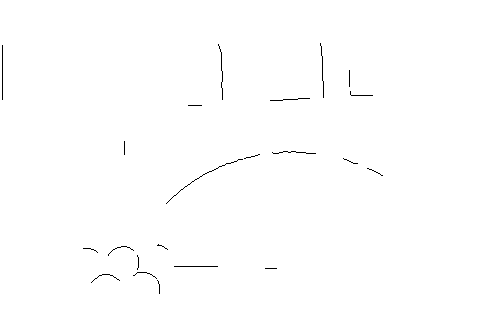}} &
 		\fbox{\includegraphics[width=0.2\textwidth]{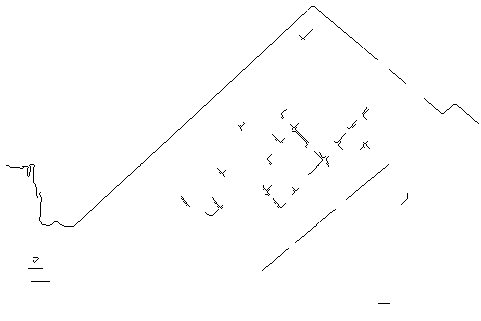}} \\	
	\end{tabular}
\caption{Visual comparison of the different boundary images obtained when applying the proposed functions in Table~\protect\ref{tab:tiposFunctions} with $S_3$ smoothing configuration (as being the best results), along with methods from the literature as $\text{Canny}$, $\text{FM}_\text{SS}$, $G_{S_P}$, $G_{S_M}$}
	\label{fig:imgBdryExtra}
\end{figure*} 

\begin{table*}
 \centering
\setlength{\tabcolsep}{3mm}
 \begin{adjustbox}{max width=0.95\textwidth}
\begin{tabular}{l c c c c c c c c c c c c }
    	\toprule
    	{} &  
				\multicolumn{3}{c}{Smoothing $S_1$} &
				\multicolumn{3}{c}{Smoothing $S_2$} &
				\multicolumn{3}{c}{Smoothing $S_3$} &
				\multicolumn{3}{c}{Smoothing $S_4$} \\
			{\small Method}&
				{\small $\text{Prec}$} & {\small $\text{Rec}$} & {\small $\text{F}_{0.5}$} &
				{\small $\text{Prec}$} & {\small $\text{Rec}$} & {\small $\text{F}_{0.5}$} &
				{\small $\text{Prec}$} & {\small $\text{Rec}$} & {\small $\text{F}_{0.5}$} &
				{\small $\text{Prec}$} & {\small $\text{Rec}$} & {\small $\text{F}_{0.5}$} \\
    	\midrule
    	\textit{{\small $\mathfrak{C}_{p^{0.8}}^{C_F}$}} & 0.625 & 0.670 & 0.622 &
				0.399 & 0.934 & 0.545 &
				0.623 & 0.744 & \textbf{0.658} &
				0.631 & 0.731 & \textbf{0.657}
				\\
			
    	\textit{{\small $\mathfrak{C}_{p^{1}}^{O_B}$}} & 0.626 & 0.674 & 0.625 &
				0.413 & 0.919 & 0.556 &
				0.627 & 0.738 & 0.657 &
				0.632 & 0.732 & \textbf{0.657}
				\\
      
			\textit{{\small $\mathfrak{C}_{p^{0.4}}^{F_{BPC}}$}} & 0.623 & 0.674 & 0.623 &
				0.398 & \textbf{0.935} & 0.545 &
				0.621 & \textbf{0.745} & \textbf{0.658} &
				0.632 & 0.731 & \textbf{0.657}
				\\
				
    	\textit{{\small $\mathfrak{C}_{p^1}^{\text{Hamacher}}$}} & 0.613 & 0.684 & 0.624 &
				0.448 & 0.861 & 0.576 &
				0.612 & 0.733 & 0.647 &
				 0.615 & \textbf{0.737} & 0.650
				\\

			\hdashline
    	
			{\small Canny}  & \textbf{0.675} & 0.647 & \textbf{0.641} &
				\textbf{0.688} & 0.617 & 0.631 &
				0.748 & 0.499 & 0.574 &
				\textbf{0.747} & 0.506 & 0.579
				\\
    	
			{\small $\text{FM}_{\text{SS}}$} 		 & 0.391 & \textbf{0.908} & 0.534 &
				0.457 & 0.797 & 0.564 &
				0.529 & 0.676 & 0.574 &
				0.556 & 0.669 & 0.583
				\\
    	
			{\small $G_{S_P}$} 		 & 0.614 & 0.667 & 0.613 &
				0.657 & 0.669 & 0.642 &
				0.742 & 0.274 & 0.369 &
				0.714 & 0.334 & 0.420
				\\
    	
			{\small $G_{S_M}$} 		 & 0.598 & 0.709 & 0.625 &
				0.647 & 0.698 & \textbf{0.651} &
				\textbf{0.757} & 0.314 & 0.413 &
				0.726 & 0.365 & 0.453
				\\
    	\bottomrule
    \end{tabular}
\end{adjustbox}
\caption{Quantitative evaluation of our method with four different smoothing configurations from Table~\protect\ref{tab:configSmoo}. The upper part of the table, over the dashed line, shows the results for the generalization of the Choquet integral functions used ($\mathfrak{C}_{p^{0.8}}^{C_F}$, $\mathfrak{C}_{p^{1}}^{O_B}$, $\mathfrak{C}_{p^{0.4}}^{F_{BPC}}$, $\mathfrak{C}_{p^1}^{\text{Hamacher}}$). The lower part of the table, under the dashed line, show the results from the literature method to whom we compare (Canny, $\text{FM}_{\text{SS}}$, $G_{S_P}$, $G_{S_M}$).}
\label{table:quantitativeEval}
\end{table*}
 
The difference obtained at a feature level can be seen in Fig.~\ref{fig:imgFeat} 
	where we show the result of the blending phase (P3) of our proposal.
	We can observe that the use of the Gaussian smoothing clearly penalizes the edge extraction process.
	Gaussian smoothing is not content-aware and makes cues to be extracted where very small intensity variation exist and hence edges detected.
	
On the contrary when the \textit{GS} approach is used,
	 as it is neighbour dependent and is capable of building homogeneous intensity regions, most of the small intensity variations are removed and less spurious elements are detected.
	 
When looking at the edges result, in Fig.~\ref{fig:imgBdry}, we can observe the effect of both smoothing processes. 
	On one hand, 
		the literature methods avoid many edges even with spurious cues present when using the Gaussian smoothing,
		except in the case of $\text{FM}_{SS}$.
	On the other hand,
		our proposal is greatly affected by the smoothing method showing that the Gaussian approach and the texture cues generate unwanted edges.
	Performing a global analysis, it is noticeable that our method performs visually better when using the \textit{GS} approach.
	With the best smoothing configuration ($S_3$) we can see in a different set of images that our four different functions perform quite equally,
		outperforming the literature methods.

In terms of quantitative analysis, 
	we show in Table~\ref{table:quantitativeEval} results related with the $(\text{Prec},\text{Rec},\text{F}_{0.5})$ measures for each smoothing configuration and for each edge processing method.
	With the gathered results we can confirm in numerical terms the results observed in the previous visual analysis.
	
    When using the Gaussian smoothing, our method (with the four different alternatives) obtains worse results than the Canny method when using $S_1$ configuration. 
	Although, it remains over the other literature methods. 
	In the case of a more intense Gaussian smoothing ($S_2$ configuration), 
		our method decays considerably and is beaten by $G_{S_M}$. 
	This results indicate that our method is subject to the influence of the blurring effect and to slight differences in intensity variations.
When using an alternative smoothing, \ie, the Gravitational one, 
	we can clearly see that our method is the best performer when using any of the generalizations of the Choquet integral,
	 getting the best result in terms of $F_{0.5}$ with $\mathfrak{C}_{p^{0.8}}^{C_F}$ and $\mathfrak{C}_{p^{1}}^{O_B}$.
If we compare the results globally, we can observe that both methods achieve a higher result than the Canny method and even $G_{S_M}$.
	 
The results when using the \textit{GS} method ($S_3$ and $S_4$) are very similar removing most of the spurious edges that are present when using any of the Gaussian smoothing configurations ($S_1$ and $S_2$).
	It is worth mentioning that the influence of the smoothing has the opposite effect on the literature methods to whom we compare, removing artifacts but also losing a great quantity pf edges.
	
When using $S_3$ configuration we have a less smoothed image, 
	getting a good compromise between regularization (with some blurring) and a good edge definition. 
With $S_4$ configuration regions are homogenized, 
	with a clearer definition in possible intensity variations (with less blurring effect)
	getting large parts of the image with very close intensity values. 
On the literature methods we can clearly see that the most strict smoothing ($S_2$) is the best approach.

More results are available in the repository at GitHub~\footnote{https://github.com/Giancarlo-Lucca/Neuro-inspired-edge-feature-fusion-using-Choquet-integrals}, where a larger number of functions for the Choquet generalizations has been put to the test along with different values for the power measure.

\section{Conclusions}\label{sec:conclusions}

In this work, we have proposed a different use of the generalizations of the Choquet integral for aggregating information in images, concretely for fusing extracted image features in the context of edge detection. With this method, we are able to represent the relationship between all the variations of intensity around each pixel, not only considering some directions, simulating in a better way the process done in the HVS.

We have analysed the behaviour of different functions for the Choquet integral generalizations, showing that the results are very similar between them and outperforms the classical methods of the literature as the Canny method.
From our experiments, the use of the Choquet integral allows for our proposal to perform better than other methods.
This fact is arguably related to the ability of the generalization of the Choquet integrals to work in terms of the relation between the elements (in this case, the image pixels).
Specifically, $\vec{r}$-monotonicity allows for the detection of edges according to the tonal direction.
In fact, in order to extract the edge cues, the relation between the pixels plays a crucial role in discriminating between true edges and outliers.

Moreover, we have tested our method with two alternative techniques for the regularization process, showing how this initial step in edge extraction process is crucial. In fact, a suitable image smoothing can make one method to be the best performer.

We can conclude that the best combination for our Choquet-like feature extraction process is to use the Gravitational smoothing, as the regularized image obtained is more homogeneous while preserving edges. This makes our method not to detect edges where small variances in the intensity information are present (\eg, textures).

As future research lines we intend to study a large variety of generalizations of the Choquet integral, like the $C_{{F_1}{F_2}}$-integrals~\cite{LUCCA2018a,gCF1F2}, which have shown promising results in Fuzzy Rule-Based Classification Systems, as well as other generalization structures such as the $d$-Choquet integrals~\cite{bustince2020d}, taking advantage of their power in comparison processes such as interval-valued data, which is also one of our future goals to improve our proposed edge detection method. Moreover, other types of fuzzy measures (\eg~weighted mean, OWA, etc.) should be studied as we have stick to the power measure, as well as using methods to learn the measure so that it better represent the relation between elements to be aggregated.

\section*{Acknowledgements}

The authors gratefully acknowledge the financial support of the Spanish Ministry of Science and Technology (project PID2019-108392GB-I00(AEI/10.13039/501100011033), the Research Services of Universidad Pública de Navarra, CNPq (307781/2016-0,	301618/2019-4), FAPERGS (19/ 2551-0001660) and PNPD/CAPES (464880/2019-00).

\bibliographystyle{./elsarticle-num}

\end{document}